\documentclass[10pt,twocolumn,letterpaper]{article}

\usepackage[pagenumbers]{cvpr} % To force page numbers, e.g. for an arXiv version

\usepackage[dvipsnames]{xcolor}

\usepackage{amsmath}
\usepackage{amssymb}
\usepackage{booktabs}
\usepackage{bm}
\usepackage{dsfont}
\usepackage{multirow}
\usepackage{makecell}
\usepackage{gensymb}
\usepackage{algorithmicx}
\usepackage{algorithm}
\usepackage{algpseudocode}
\usepackage{xcolor}
\usepackage{scrextend}
\makeatletter
\@namedef{ver@everyshi.sty}{}
\makeatother
\usepackage{tikz}
\usepackage{amsthm}

\definecolor{cvprblue}{rgb}{0.58,0.3,0.0}
\usepackage[pagebackref,breaklinks,colorlinks,citecolor=cvprblue]{hyperref}

\usepackage[capitalize]{cleveref}
\crefname{section}{Sec.}{Secs.}
\Crefname{section}{Section}{Sections}
\Crefname{table}{Table}{Tables}
\crefname{table}{Tab.}{Tabs.}
\crefname{Lemma}{Lemma.}{Lemmas.}

\def\paperID{xxxx} % *** Enter the Paper ID here
\def\confName{xxxx}
\def\confYear{2024}

\title{Sur$^{2}$f: A Hybrid Representation for High-Quality and Efficient Surface Reconstruction from Multi-view Images}

\begin{document}
%%%%%%%%% AUTHORS - PLEASE UPDATE
\author{
    Zhangjin Huang\textsuperscript{1,*},
    Zhihao Liang\textsuperscript{1,*},
    Haojie Zhang\textsuperscript{1},
    Yangkai Lin\textsuperscript{1}, 
    Kui Jia\textsuperscript{2,\textdagger} \\
    \textsuperscript{1}South China University of Technology \\
    \textsuperscript{2}School of Data Science, The Chinese University of Hong Kong, Shenzhen\\
    {\tt\small \{eehuangzhangjin,eezhihaoliang\}@mail.scut.edu.cn},
    {\tt\small kuijia@cuhk.edu.cn}\\
    \tt\small{\url{https://huang-zhangjin.github.io/project-pages/sur2f.html}}
}
\maketitle

\let\thefootnote\relax\footnote{* indicates equal contribution.}

\let\thefootnote\relax\footnote{\textsuperscript{\textdagger}Correspondence to Kui Jia $<$kuijia@cuhk.edu.cn$>$.}

%%%%%%%%%%%%%%%%%%%%%%%%%%%%%%%%%%%%%%%%%%%%%%%%%%%%%%%%%%%%%%%%%%%%%%%%%%%%%%%%%%%%%%%%%%%%%%%%%%%
\vspace{-1cm}
\begin{abstract}
Multi-view surface reconstruction is an ill-posed, inverse problem in 3D vision research. It involves modeling the geometry and appearance with appropriate surface representations. Most of the existing methods rely either on explicit meshes, using surface rendering of meshes for reconstruction, or on implicit field functions, using volume rendering of the fields for reconstruction. The two types of representations in fact have their respective merits. In this work, we propose a new hybrid representation, termed Sur$^2$f, aiming to better benefit from both representations in a complementary manner. Technically, we learn two parallel streams of an implicit signed distance field and an explicit surrogate surface (Sur$^2$f) mesh, and unify volume rendering of the implicit signed distance function (SDF) and surface rendering of the surrogate mesh with a shared, neural shader; the unified shading promotes their convergence to the same, underlying surface. We synchronize learning of the surrogate mesh by driving its deformation with functions induced from the implicit SDF. In addition, the synchronized surrogate mesh enables surface-guided volume sampling, which greatly improves the sampling efficiency per ray in volume rendering. We conduct thorough experiments showing that Sur$^2$f outperforms existing reconstruction methods and surface representations, including hybrid ones, in terms of both recovery quality and recovery efficiency.
\end{abstract}

\begin{figure}
\vspace{-0.1cm}
\centering
\scalebox{0.45}{
    \includegraphics[width=1.0\textwidth]{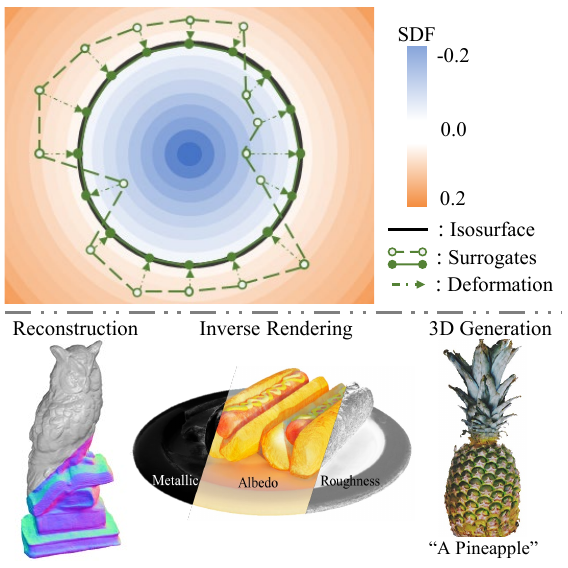}
}
\vspace{-0.3cm}
\caption{
    The proposed Sur$^2$f and its applications. 
}
\label{fig:teaser}
\vspace{-0.4cm}
\end{figure}

\section{Introduction}
\label{sec:intro}
Surface reconstruction of 3D objects or scenes from a set of multi-view observed images is a classical and challenging problem in 3D vision research. The problem is ill-posed given that multi-view images are only 2D observations of the 3D contents. To achieve surface reconstruction, recent efforts \cite{chen2019learning,worchel2022multi,goel2022differentiable,yariv2020multiview,oechsle2021unisurf,yariv2021volume,wang2021neus} aim to build the inverse process of image projection or rendering, such that the underlying geometry and surface appearance can be recovered. The process involves modeling the underlying geometry and appearance for inverse rendering, for which different surface representations can be adopted. For example, one may represent the surface \emph{explicitly} as a polygonal mesh and use an associated shading function for surface rendering; alternatively, one may model the surface geometry as (level set of) an \emph{implicit} field (\eg, signed distance function (SDF) \cite{park2019deepsdf} or occupancy/opacity field \cite{mescheder2019occupancy,mildenhall2020nerf}), and render images of different views via volume rendering \cite{mildenhall2020nerf}. By parameterizing either explicit or implicit surface representations and developing differentiable versions of the corresponding surface or volume rendering techniques \cite{liu2019soft,Laine2020diffrast,mildenhall2020nerf,yariv2021volume,wang2021neus}, the underlying surface can be recovered.
% the underlying surface captured by the images can be recovered. 

The explicit representation of a triangle mesh is compatible with the existing graphics pipeline, and it supports efficient rendering (\eg, via rasterization or ray casting); however, directly optimizing the mesh parameterization (\eg, positions of vertices, edge connections, etc.) in the context of inverse rendering is less convenient, since complex topologies are difficult to be obtained in the optimization process. In contrast, an implicit representation models the geometry by learning in the continuous 3D space, and consequently, it supports smooth optimization of model parameters and geometry of complex topologies can be recovered by isosurface extraction from the learned field function; however, learning via volume rendering of an implicit field is of low efficiency and less precise in terms of recovering the sharp surface \cite{mildenhall2020nerf}. Given these pros and cons of different representations, a few hybrid representations \cite{shen2021deep_dmtet,shen2023flexible} have been proposed aiming for better leveraging of their respective advantages. Notably, DMTet \cite{shen2021deep_dmtet} proposes to learn a signed distance function that controls the deformation of a parameterized tetrahedron, and with a differentiable marching tetrahedra layer for mesh extraction, it supports efficient and differentiable surface rendering that enables learning from the observed multi-view images. DMTet is improved by FlexiCubes \cite{shen2023flexible} and is used in NVDIFFREC \cite{nvdiffrec} for jointly optimizing shape, material, and lighting from multi-view images. However, implicit functions used in DMTet and its variants are discretely defined on tetrahedral grid vertices, and as such, the key benefits of optimization in continuous 3D spaces from implicit field representations are only enjoyed partially; in addition, they do not support rendering from its implicit functions and thus cannot directly receive supervisions from the image observations.

In this work, we propose a new hybrid representation that aims to make a better use of the explicit and implicit surface representations. Technically, we learn two parallel streams of an implicit SDF and an explicit, \emph{surrogate surface (Sur$^2$f)} mesh; Fig.\ref{fig:pipeline} gives an illustration. We term our proposed representation as Sur$^2$f, given that this surrogate plays an essential role in the learning. More specifically, we enforce learning of the pair of representations to be synchronized by driving the deformation of the surrogate mesh with functions induced from the implicit SDF. 
%{\color{red} We prove that under mild conditions, the deformation is guaranteed to converge to a synchronization.}
For the implicit stream, we use SDF-induced volume rendering \cite{yariv2021volume,wang2021neus} to receive supervision from images of view observations. For the stream of explicit surrogate, we use efficient and differentiable surface rendering to receive additional supervision. We unify volume rendering and surface rendering of the two streams with a shared, neural shader, which is parameterized to support photorealistic rendering. Sharing the neural shader between the two representations is crucial to our proposed Sur$^2$f, as it promotes the two parallelly learned representations to converge to the same surface. Maintaining the synchronized surrogate surface mesh also brings an important benefit for volume rendering --- it improves the sampling efficiency per ray greatly, which consequently improves the quality of volume rendering.   
To verify our proposed Sur$^2$f, we conduct thorough experiments in the contexts of inverse rendering and multi-view surface reconstruction. Experiments show that in terms of both recovery quality and recovery efficiency, our method is better than both the existing hybrid representations and the existing methods of multi-view surface reconstruction. As a general hybrid representation, we show the usefulness of Sur$^2$f for other surface modeling and reconstruction tasks as well. 
We summarize our technical contributions as follows. 
\begin{itemize}
\item We propose a new hybrid representation, termed Sur$^2$f, that can enjoy the benefits of both explicit and implicit surface representations. This is achieved by learning two parallel streams of an implicit SDF and an explicit surrogate surface mesh, both of which, by rendering, receive supervision from multi-view image observations.
\item We unify volume rendering of the implicit SDF and surface rendering of the surrogate mesh with a shared, neural shader, which promotes the two parallelly learned representations to converge to the same surface.
\item Learning of the surrogate mesh is synchronized by driving its deformation with functions induced from the implicit SDF. The surrogate mesh also enables surface-guided volume sampling, which greatly improves the sampling efficiency per ray in volume rendering.  
\item We conduct thorough experiments in various task settings of surface reconstruction from multi-view images. Our proposed Sur$^2$f outperforms existing surface representations and reconstruction methods in terms of both recovery quality and recovery efficiency.   
\end{itemize}

\section{Related Works}
\label{sec:relatedworks}

In this section, we briefly review the literature of explicit and implicit surface representations, and how they can be respectively rendered in a differentiable manner. We also review representative methods of hybrid representations. Some of these methods are state-of-the-art for the tasks of inverse rendering and multi-view surface reconstruction.

\vspace{0.1cm}
\noindent\textbf{Differentiable rendering of explicit meshes} 
% {\color{red} XXX please briefly review works of different rasterization \cite{liu2019soft,Laine2020diffrast}, emphasizing its efficiency and compabiblity with classic graphics, and the work \cite{worchel2022multi}, any others? XXX}
Traditional graphics pipelines (\eg OpenGL \cite{shreiner2009opengl}) offer hardware-accelerated rendering capabilities, however, they incorporate a discretization phase of rasterizing polygonal meshes that prevents the rendering process from being differentiable.
Recent efforts \cite{liu2019soft,chen2019learning,Laine2020diffrast} develop differentiable versions of the rasterization and facilitate studies \cite{goel2022differentiable,worchel2022multi,walker2023explicit,fastmesh} to infer 3D geometry from 2D images. Specifically, these studies optimize mesh vertices through gradient backpropagated from the rendering process, yet the optimization of complex topologies remains an unresolved problem. 

\vspace{0.1cm}
\noindent\textbf{Volume rendering of implicit fields} Popular implicit representations include SDF \cite{park2019deepsdf}, occupancy field \cite{mescheder2019occupancy}, and density/opacity field \cite{mildenhall2020nerf}.
For rendering, early surface rendering methods \cite{niemeyer2020differentiable,liu2020dist,yariv2020multiview} seek ray intersection with the underlying surface represented by SDF or occupancy field and estimate shading color with a neural network. Recent volumetric rendering methods \cite{mildenhall2020nerf,muller2022instant,TensoRF} marching rays through the whole space that are represented by density/opacity field for rendering. NeuS \cite{wang2021neus} and VolSDF \cite{yariv2021volume} provide conversions from SDF into density for volume rendering. 
And their follow-up works \cite{wang2023neus2,neuralangelo,wu2022voxurf,li2022voxsurf,Pet-Neus,fu2022geo,cai2023neuda,NeuralWarp,HF-Neus} make efforts to improve the quality and/or efficiency.
Notably, UNISURF \cite{oechsle2021unisurf} gradually reduces sampling region to encourage volume rendering to surface rendering.
% {\color{red} For rendering,  XXX please review the most important works from \cite{wang2021neus,yariv2021volume,yariv2020multiview,liu2020dist,niemeyer2020differentiable} XXX.}

\vspace{0.1cm}
\noindent\textbf{Hybrid surface representations} 
% {\color{red} XXX please review works of DMTet \cite{shen2021deep_dmtet}, FlexiCubes \cite{shen2023flexible}, and NVDIFFREC \cite{nvdiffrec}. Since we already mention them in Section 1, the review can be made very briefly XXX. }
There also exist some hybrid surface representations in the literature. 
DMTet \cite{shen2021deep_dmtet} and its subsequent work FlexiCubes \cite{shen2023flexible} represent a surface with explicit grid and implicit SDF, where mesh could be extracted by differentiable marching operation.
They are adopted by NVDIFFREC \cite{nvdiffrec} for physically based inverse rendering from multi-view images.

\section{Preliminaries on rendering of different surface representations}
\label{sec:preliminary}
In this section, we present representatives of existing rendering techniques for either implicit or explicit surface representations, which also prepare for the presentation of our proposed, new hybrid representation. 

\vspace{0.1cm}
\noindent\textbf{Rendering of an implicit SDF}
An SDF-based implicit representation can be either volume-rendered \cite{yariv2021volume,wang2021neus}, following NeRF \cite{mildenhall2020nerf}, or directly rendered on the surface \cite{yariv2020multiview}. For volume rendering of an SDF, we denote 
%Differentiable volume rendering is used in NeRF \cite{mildenhall2020nerf} for the task of novel view synthesis.
a ray emanating from the camera center and through the rendering pixel as $\bm{r}(t) = \bm{o} + t\bm{\omega}$, $t\geq 0$, where $\bm{o}\in\mathbb{R}^3$ is the camera center and $\bm{\omega}\in\mathbb{R}^3$, $\Vert \bm{\omega} \Vert = 1$, denotes the unit vector of viewing direction. 
Assume $N$ points are sampled along $\bm{r}$ and we write the $i$-th sampled point as $\bm{x}_i = \bm{r}(t_i) = \bm{o} + t_i \bm{\omega}$; the shading color accumulated along the ray $\bm{r}$ can be approximated using the quadrature rule \cite{max1995optical} as:
\vspace{-0.15cm}
\begin{equation}
\label{eq:volume_render}
\begin{aligned}
    &\bm{C}_v(\bm{r}) = \sum^N_{i=1} T_i \alpha_i \bm{c}_v(\bm{x}_i, \bm{n}_i, \bm{\omega}), \\
    &T_i = \prod^{i - 1}_{j = 1}(1 - \alpha_j), \quad
     \alpha_i = 1 - \exp(-\tau(\bm{x}_i) \delta_i),
\end{aligned}
\end{equation}
where $\delta_i = t_{i + 1} - t_i$ denotes the distance between adjacent samples. 
While NeRF uses an MLP to directly encode the volume density $\tau : \mathbb{R}^3 \rightarrow \mathbb{R}_+$, VolSDF \cite{yariv2021volume} and NeuS \cite{wang2021neus} further show that $\tau$ can be modeled as a transformed function of the implicit SDF $f: \mathbb{R}^3 \rightarrow \mathbb{R}$ \footnote{For any point $\bm{x} \in \mathbb{R}^3$ in a 3D space, $| f(\bm{x}) |$ assigns its distance to the surface $\mathcal{S}=\{\bm{x} \in \mathbb{R}^3 | f(\bm{x})=0\}$; by convention, we have $f(\bm{x}) < 0$ for points inside the surface and $f(\bm{x}) > 0$ for those outside.}, enabling better recovery of the underlying geometry. An SDF can also be used for neural surface rendering. For example, in \cite{yariv2020multiview}, 
the intersection of the ray and the surface represented by an SDF is obtained by the sphere tracing algorithm \cite{hart1996sphere} and is then fed into a neural shader to estimate the shading color.
% {\color{red} XXX use a few sentences to describe how neural surface rendering in \cite{yariv2020multiview} is enabled given an SDF, by learning a neural shader that approximates PBR XXX}

\vspace{0.1cm}
\noindent\textbf{Rendering of an explicit surface}
An explicit surface mesh can be rendered with better consideration of physical properties, \ie, physics-based rendering (PBR). 
%In contrast to volume rendering, where pixel shading is integrated over the whole corresponding ray, surface rendering just considers the intersection point between the ray and surface.
Denote a triangle mesh as $\mathcal{S} = \{\mathcal{V}, \mathcal{F}\}$, which is composed of a set $\mathcal{V}$ of vertices $\{ \bm{v} \in \mathbb{R}^3 \}$ and a set $\mathcal{F}$ of faces. Assume that a ray $\bm{r}$ intersects $\mathcal{S}$ at the surface point $\bm{x}_{\mathcal{S}} = \bm{r}(t_{\mathcal{S}})$, and we denote the surface normal as $\bm{n}_{\bm{x}_{\mathcal{S}}}$. 
The shading for the surface point $\bm{x}_{\mathcal{S}}$ from a camera view $\bm{{\omega}}$ is formulated as the non-emissive rendering equation \cite{kajiya1986rendering}:
\vspace{-0.15cm}
\begin{equation}
    \label{eq:surface_rendering}
    L(\bm{x}_{\mathcal{S}}, \bm{\omega}) = \int_\Omega L_i(\bm{x}_{\mathcal{S}}, \bm{l})B(\bm{l}, - \bm{\omega}) \left<\bm{l} \cdot \bm{n}_{\bm{x}_{\mathcal{S}}} \right> d\bm{l}, 
\end{equation}
where $\Omega$ denotes the hemisphere centered at $\bm{x}_{\mathcal{S}}$ and rotating around $\bm{n}_{\bm{x}_{\mathcal{S}}}$, $\bm{l} \in\mathbb{R}^3$, $\Vert \bm{l} \Vert = 1$, denotes an incoming light direction, $L_i(\bm{x}, \bm{l})$ denotes the incoming radiance of $\bm{x}$ from $\bm{l}$, $\left<\cdot\right>$ is a descriptor that computes the cosine of two input vectors, and $B$ is the bidirectional reflectance distribution function (BRDF) that describes the portion of reflected radiance at direction $-\bm{\omega}$ from direction $\bm{l}$.
Based on the basic rendering of \cref{eq:surface_rendering}, \cite{yariv2020multiview,worchel2022multi} use MLP-based neural shaders to approximate PBR; more specifically, a coordinate-based MLP takes the intersection point $\bm{x}_{\mathcal{S}}$, its associated surface normal $\bm{n}_{\bm{x}_{\mathcal{S}}}$, and the view direction $\bm{\omega}$ as inputs, and learns to output the shading result as:
\vspace{-0.15cm}
\begin{equation}
\label{eq:surface}
    \bm{C}_s(\bm{r}) = \bm{c}_s(\bm{x}_{\mathcal{S}}, \bm{n}_{\bm{x}_{\mathcal{S}}}, \bm{\omega}).
\end{equation}

\begin{figure*}
\centering
\scalebox{0.96}{
    \includegraphics[width=1.0\textwidth]{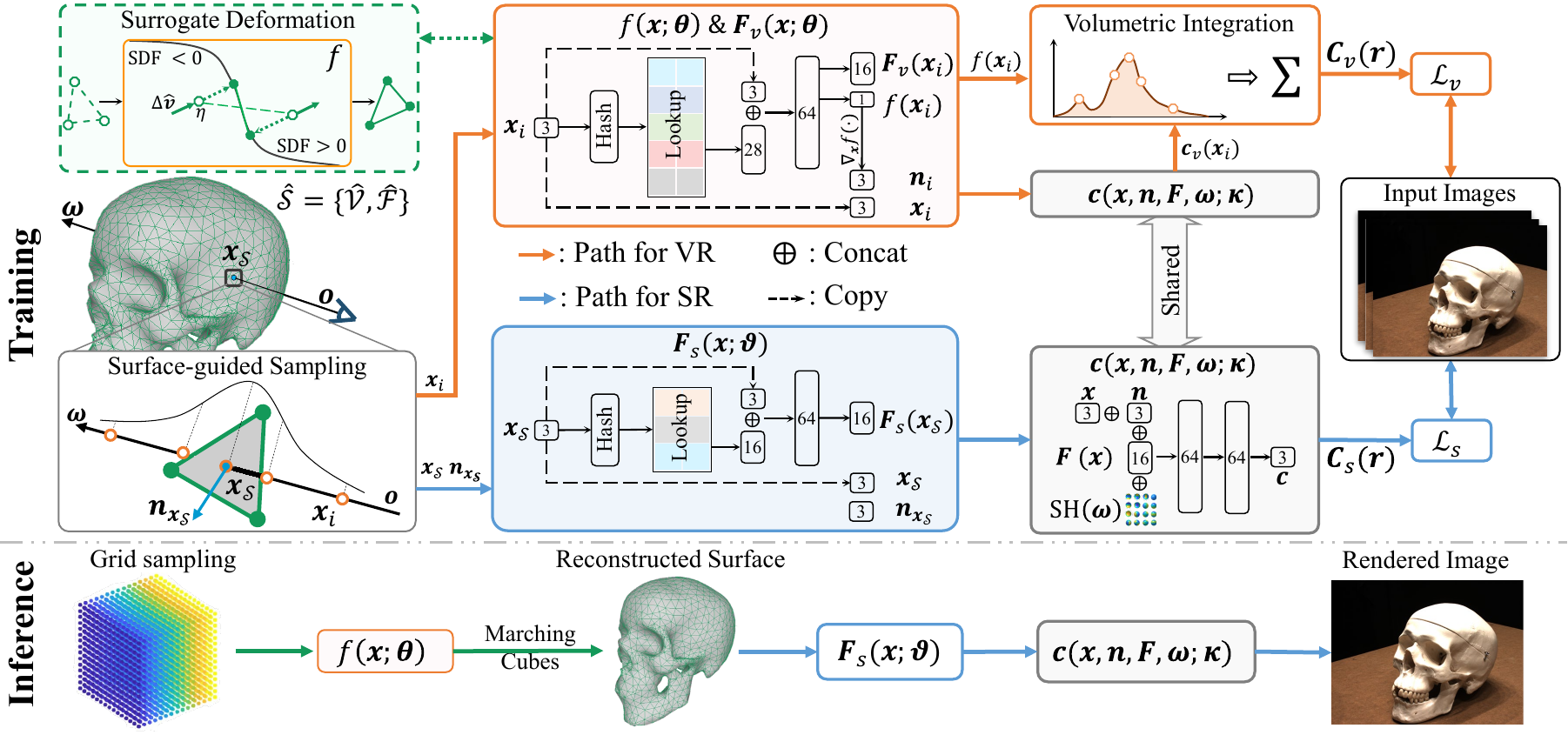}
}
    \vspace{-0.2cm}
    \caption{\textbf{Overview of Sur$^2$f}.  Sur$^2$f learns parallel streams of an implicit SDF $f$ and a surrogate surface mesh $\widehat{\mathcal{S}} = \{ \widehat{\mathcal{V}}, \widehat{\mathcal{F}} \}$. Deformation of vertices $\{ \widehat{\mathcal{V}} \}$ of the surrogate $\widehat{\mathcal{S}}$ is driven by functions induced from $f$, which enforces synchronization of the two representations (\cf Section \ref{subsec:surrogate}). We use a shared, neural shader for both SDF-induced volume rendering and surface rendering of $\widehat{\mathcal{S}}$ (\cf Section \ref{subsec:commonshading}), which enables dual supervisions from image observations. We also use surface-guided volume sampling (\ie, guided by $\widehat{\mathcal{S}}$) to improve the sampling efficiency (and consequently reconstruction quality) of volume rendering (\cf Section \ref{subsec:sampling}). During inference, we use marching cubes \cite{lorensen1987marching} to extract a mesh $\mathcal{S}$ from $f$; with the already learned neural shader, we achieve photorealisc, real-time rendering from Sur$^2$f.  
    }
\label{fig:pipeline}
\vspace{-0.3cm}
\end{figure*}

%-------------------------------------------------------------------------
\section{The proposed hybrid representation}
\label{sec:method}

We present our proposed hybrid representation in the context of surface reconstruction from multi-view image observations. Given a set of calibrated RGB images $\{\bm{I}_m\}^M_{m=1}$ of a 3D scene, the task is to reconstruct the scene surface $\mathcal{S}$ and the associated surface appearance.
Under the framework of differentiable rendering, the task boils down as modeling and learning the appropriate surface representations that are rendered to match $\{\bm{I}_m\}^M_{m=1}$. 
As discussed in Sections \ref{sec:intro} and \ref{sec:preliminary}, one may rely either on volume rendering of an implicit field, or on surface rendering of an explicit mesh; each of them has its own merits, while being limited in other aspects. Simply to say, volume rendering of implicit fields is easier to be optimized and has the flexibility in learning complex typologies; it is, however, of low efficiency and less compatibility with existing graphics pipelines. In contrast, surface rendering of explicit meshes has the opposite pros and cons. 
We are thus motivated to propose a new, hybrid representation that can benefit from both representations. We technically achieve the goal by learning a synchronized pair of an implicit SDF $f$ and an explicit \emph{surrogate surface (Sur$^2$f)} mesh $\widehat{\mathcal{S}} = \{ \widehat{\mathcal{V}}, \widehat{\mathcal{F}} \}$; we enforce their synchronization by driving the iterative deformation of $\widehat{\mathcal{S}}$ with functions induced from the implicit $f$ that is optimized to the current iteration. We further unify SDF-induced volume rendering of $f$ and surface rendering of $\widehat{\mathcal{S}}$ with a shared, neural shader, where the synchronized $\widehat{\mathcal{S}}$ is also used to greatly improve the sampling efficiency in volume rendering. Fig. \ref{fig:pipeline} gives an illustration. 
Compared with the hybrid representation of DMTet \cite{shen2021deep_dmtet}, the implicit part of Sur$^2$f is continuously defined and can be directly rendered to receive supervision from $\{\bm{I}_m\}^M_{m=1}$. Consequently, Sur$^2$f is expected to be better in terms of surface modeling and reconstruction. 
In Section \ref{sec:experiments}, we validate our proposed Sur$^2$f for experiments of multi-view surface reconstruction as well as other surface modeling tasks.
% In Section \ref{sec:experiments}, we validate our proposed Sur$^2$f for experiments of multi-view surface reconstruction. As a general hybrid representation, Sur$^2$f is also useful for other surface modeling tasks as shown in Section \ref{sec:experiments}.
% As a general hybrid representation, we show in Section \ref{sec:experiments} that Sur$^2$f is useful for other surface modeling and reconstruction tasks as well. 

\subsection{SDF-induced volume rendering}
\label{subsec:neus}

As shown in the top, orange path in Fig. \ref{fig:pipeline}, we learn a neural implicit function of SDF $f(\cdot;\bm{\theta}) \in \mathbb{R}$ whose zero-level set represents the underlying surface $\mathcal{S}=\{\bm{x} \in \mathbb{R}^3 | f(\bm{x})=0\}$ to be reconstructed; $f(\cdot;\bm{\theta})$ takes any space point $\bm{x} \in \mathbb{R}^3$ as input and returns its signed distance to $\mathcal{S}$.   
In this work, we implement $f(\cdot;\bm{\theta})$ as an MLP and encode any $\bm{x}$ using multi-resolution hash grid \cite{muller2022instant}, before feeding into the network.  
For any space point $\bm{x}$, except for the signed distance value $f(\cdot;\bm{\theta})$, we also learn its feature as $\bm{F}_v(\bm{x}; \bm{\theta})$, using the same MLP network, as shown in Fig. \ref{fig:pipeline}; note that both $f(\cdot;\bm{\theta})$ and $\bm{F}_v(\bm{x}; \bm{\theta})$ are parameterized by the same $\bm{\theta}$. 

% Given $f$, we conduct SDF-induced volume rendering, following NeuS \cite{wang2021neus}. We use the unbiased volume rendering \cite{wang2021neus} and the discrete opacity $\alpha_i$ in \cref{eq:volume_render} is described as:
Given $f$, we conduct SDF-induced volume rendering, following NeuS \cite{wang2021neus}, where the discrete opacity $\alpha_i$ in \cref{eq:volume_render} is described as:
\begin{equation}
\label{eq:neus}
    \alpha_i = \max\left(\frac{\Phi_s(f(\bm{x}_i)) - \Phi_s(f(\bm{x}_{i + 1}))}{\Phi_s(f(\bm{x}_i))}, 0\right),
\end{equation}
% where $\Phi_s(z) = 1 / (1 + e^{-sz})$ is the Sigmoid function, {\color{red} $s$ is a scaling hyper-parameter}, and $\bm{x}_i = \bm{r}(t_i) = \bm{o} + t_i \bm{\omega}$ is a sampled point along a ray $\bm{r}$. {\color{red} We also note that a normal $\bm{n}_i$ at $\bm{x}_i$ can be computed as XXX. }  
where $\Phi_s(z) = 1 / (1 + e^{-sz})$ is the Sigmoid function, $s$ is a learnable parameter, and $\bm{x}_i = \bm{r}(t_i) = \bm{o} + t_i \bm{\omega}$ is a sampled point along a ray $\bm{r}$. We also note that a normal $\bm{n}_i$ at $\bm{x}_i$ can be computed as the derivation of $f$: $\bm{n}_i = \nabla_{\bm{x}_i} f(\bm{x}_i)$.

\subsection{Learning a synchronized surrogate of explicit mesh}
\label{subsec:surrogate}

A useful hybrid surface representation expects an explicit mesh whose deformation is synchronized with (zero-level set of) the implicit $f$. This can be simply achieved by applying isosurface extraction per iteration from the updated $f$, as in \cite{shen2021deep_dmtet,shen2023flexible}. 
However, isosurface extraction per iteration is computationally expensive, and more importantly, for a hybrid representation, it may not provide complementary benefits on the surface learning, since the extracted mesh is equivalent to the implicit field updated per iteration.  

\begin{figure}%[htbp]
\centering
\scalebox{0.46}{
    \includegraphics[width=\textwidth]{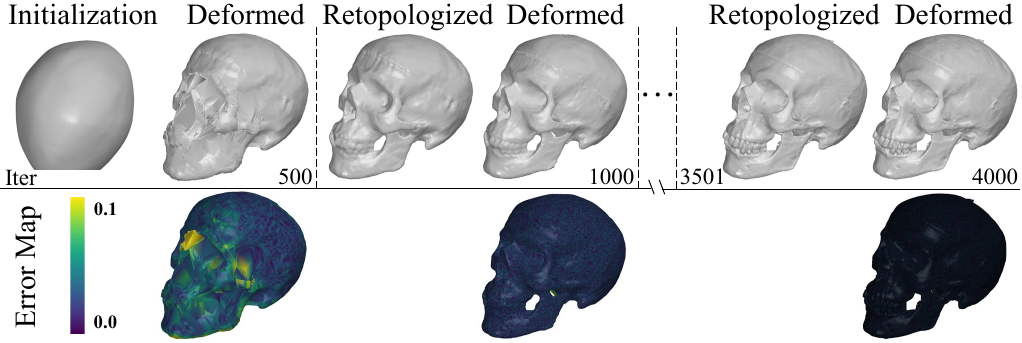}
}
% \resizebox{0.76\linewidth}{!}{\begin{tabular}{p{70pt}<{\centering}p{70pt}<{\centering}p{75pt}<{\centering}p{75pt}<{\raggedright}p{70pt}<{\centering}p{70pt}} Initialization & Deformed & Retopologized & Deformed & Retopologized & Deformed  \end{tabular}}
% \begin{minipage}{0.46\textwidth}
%     \centering
%     \includegraphics[width=\textwidth]{images/main/ablfig_syndeform1.pdf}
% \end{minipage}
% \begin{minipage}{0.04\textwidth}
% \resizebox{0.03\textwidth}{!}{\rotatebox[origin=c]{90}{Error Map}}
% \end{minipage}
% \begin{minipage}{0.46\textwidth}
%     \centering
%     \includegraphics[width=\textwidth]{images/main/ablfig_syndeform2.pdf}
% \end{minipage}
\vspace{-0.3cm}
    \caption{Visualization of the synchronized deformation of the surrogate mesh.  
    Each error map shows the distance between the surrogate mesh and the mesh extracted from the SDF $f$ at a certain training iteration.
    }
    \label{fig:ablfig_synsurrogate}
\vspace{-0.4cm}
\end{figure}

In this work, we propose to maintain an explicit, surrogate surface mesh $\widehat{\mathcal{S}} = \{ \widehat{\mathcal{V}}, \widehat{\mathcal{F}} \}$ at a much lower computational cost, whose iterative deformation is driven by, but not directly extracted from, $f$. More specifically, we deform each vertex $\hat{\bm{v}} \in \widehat{\mathcal{V}}$ per iteration according to:  
\begin{equation}
\label{eq:deform}
\begin{aligned}
    & \hat{\bm{v}} \leftarrow \hat{\bm{v}} - \eta\Delta\hat{\bm{v}}, \\
    \textrm{with} \quad \eta = f(\hat{\bm{v}}), & \quad \Delta\hat{\bm{v}} = \nabla_{\hat{\bm{v}}} f(\hat{\bm{v}}) / \left\Vert \nabla_{\hat{\bm{v}}} f(\hat{\bm{v}}) \right\Vert_2 .
\end{aligned}
\end{equation}
An initial $\widehat{\mathcal{S}}$ can be obtained by applying marching cubes \cite{lorensen1987marching} to an SDF $f$ whose zero-level set is assumed to capture an initial surface (\eg, a sphere). We conduct empirical analysis on the synchronization via rule (\cref{eq:deform}) between $\widehat{\mathcal{S}}$ and $f$ (\cf Section \ref{sec:experiments} for details of the empirical setting). Fig. \ref{fig:ablfig_synsurrogate} shows that when the training of Sur$^2$f proceeds, the difference between $\widehat{\mathcal{S}}$ and the zero-level set of $f$ is gradually reduced. Indeed, training improves the conditioning of the level sets of $f$, and in the ideal case, we have $\nabla_{\bm{x}_i}f(\bm{x}_i) \sim \nabla_{\bm{x}_j}f(\bm{x}_j) \sim \nabla_{\bm{x}_0}f(\bm{x}_0)$ for any pair of $\bm{x}_i$ and $\bm{x}_j$ that are on different level sets of $f$,  \ie, $f(\bm{x}_i) \neq f(\bm{x}_j)$, but correspond to the same surface point $\bm{x}_0$, where $f(\bm{x}_0) = 0$; in other words, normals at these space points are parallel. When coming to this ideal case, we have $f\big( \bm{x} - f(\bm{x}) \nabla_{\bm{x}} f(\bm{x}) / \left\Vert \nabla_{\bm{x}} f(\bm{x}) \right\Vert_2 \big) = 0$, which means that a single-step updating by (\cref{eq:deform}) synchronizes the two representations. In practice, we observe approaching of this ideal case at later stages of training, at least for space points near the surface. 
% Considering that the deformation (\cref{eq:deform}) itself is not guaranteed to meshes of complex topologies, we periodically apply marching cubes to re-boot the process; ablation study in Section \ref{sec:experiments} shows that re-booting every 500 iterations is enough to produce high-quality meshes, while keeping the deformation of high efficiency. 
Considering that the deformation (\cref{eq:deform}) itself is not guaranteed to meshes of complex topologies, we periodically apply marching cubes to re-boot the process; ablation study in Section \ref{sec:experiments} shows that re-booting every 500 iterations is enough to produce high-quality meshes while keeping a high efficiency.

%We show the convergence of the rule (\cref{eq:deform}) to the zero-level set of $f$, assuming that $f$ has learned to capture a well-defined surface mesh. 

%Lemma \ref{Lemma:lemma1} shows that when $f$ is optimized to the underlying true surface $\mathcal{S}$, our used updating rule (\cref{eq:deform}) is guaranteed to synchronize the deformation of $\widehat{\mathcal{S}}$ towards $\mathcal{S}$. In practice, $f$ is iteratively updated and during the optimization process, we empirically observe that the surrogate $\widehat{\mathcal{S}}$ is deformed w.r.t. the zero-level set of the currently updated $f$, and they together converge to an almost identical surface, which is the final result of the learning process.
% {\color{red} we empirically observe that the surrogate $\widehat{\mathcal{S}}$ is deformed w.r.t. the zero-level set of the currently updated $f$, and they together converge to an almost identical surface, which is the final result of the learning process. --- am I correct? Do we have the experiments to support? --- }
%To initialize $\widehat{\mathcal{S}}$, we use marching cubes \cite{lorensen1987marching} to extract from $f$, and we periodically apply marching cubes to re-boot the process, considering that the deformation (\cref{eq:deform}) itself cannot produce meshes of complex topologies; ablation study in Section \ref{sec:experiments} shows that re-booting every 500 iterations is enough to produce high-quality meshes, while keeping the deformation of high efficiency. 

\subsection{Unifying surface rendering and volume rendering with a shared shading network}
\label{subsec:commonshading}

Given the surrogate surface $\widehat{\mathcal{S}}$, a surface point $\bm{x}_{\widehat{\mathcal{S}}} = \bm{r}(t_{\widehat{\mathcal{S}}}) = \bm{o} + t_{\widehat{\mathcal{S}}} \bm{\omega}$ can be efficiently and exactly obtained in a device-native manner (\eg, via rasterization or ray casting), which is an intersection of the ray $\bm{r}$ and $\widehat{\mathcal{S}}$. For any $\bm{x}_{\widehat{\mathcal{S}}}$, we use an MLP, parameterized by $\bm{\vartheta}$, to learn its geometric feature as $\bm{F}_s(\bm{x}_{\widehat{\mathcal{S}}}; \bm{\vartheta})$, and we again encode $\bm{x}_{\widehat{\mathcal{S}}}$ using multi-resolution hash grid \cite{muller2022instant}, before feeding into the MLP network. 
% {\color{red} The surface normal $\bm{n}_{\widehat{\mathcal{S}}}$ at $\bm{x}_{\widehat{\mathcal{S}}}$ is computed as XXX. } 
The surface normal $\bm{n}_{\widehat{\mathcal{S}}}$ at $\bm{x}_{\widehat{\mathcal{S}}}$ is the normal vector associated with the intersecting triangle face \cite{botsch2010polygon}.
The bottom, blue path in Fig. \ref{fig:pipeline} gives the illustration.   

Comparing the shading function $\bm{c}_v(\bm{x}_i, \bm{n}_i, \bm{\omega})$ in (\cref{eq:volume_render})  for volume rendering $\bm{C}_v(\bm{r})$ of a ray $\bm{r}$ and its counterpart $\bm{c}_s(\bm{x}_{\mathcal{S}}, \bm{n}_{\mathcal{S}}, \bm{\omega})$ in (\cref{eq:surface}) for surface rendering $\bm{C}_s(\bm{r})$ at the ray-surface intersection, one may be tempted to unify them using a same shading function. In fact, given appropriate hypothesis spaces of functions, both shading functions can model PBR \cite{yariv2020multiview}. The difference lies in that volume rendering of (\cref{eq:volume_render}) weights and sums the shadings for $\{ \bm{x}_i \}$ sampled along $\bm{r}$, while surface rendering of (\cref{eq:surface}) concentrates sharply on the surface point; as such, the two ways of shading converge to an identical one when the distribution of weights in (\cref{eq:volume_render}) (\ie, $\{ T_i \alpha_i \}$) approaches to a Dirac one. In this work, we propose surface-guided volume sampling (\cf Section \ref{subsec:sampling}) that practically enhances this effect.   
Also considering that an MLP of enough capacity is a universal function approximator, we propose to unify $\bm{c}_v(\bm{x}_i, \bm{n}_i, \bm{\omega})$ and $\bm{c}_s(\bm{x}_{\mathcal{S}}, \bm{n}_{\mathcal{S}}, \bm{\omega})$ using a shared neural shader $\bm{c}(\bm{x}, \bm{n}, \bm{F},\bm{\omega}; \bm{\kappa})$, parameterized by $\bm{\kappa}$, where we also augment with a feature $\bm{F}$ that encodes the local geometry around $\bm{x}$. $\bm{c}(\bm{x}, \bm{n}, \bm{F},\bm{\omega}; \bm{\kappa})$ is instantiated as $\bm{c}(\bm{x}_i, \bm{n}_i, \bm{F}_v(\bm{x}_i), \bm{\omega}; \bm{\kappa})$ for volume rendering of sampled $\{\bm{x}_i\}$ along a ray, and as $\bm{c}(\bm{x}_{\widehat{\mathcal{S}}}, \bm{n}_{\bm{x}_{\widehat{\mathcal{S}}}}, \bm{F}_s(\bm{x}_{\widehat{\mathcal{S}}}), \bm{\omega}; \bm{\kappa})$ for rendering of a surface point on the surrogate $\widehat{\mathcal{S}}$. In practice, we also use sphere harmonics \cite{yu2021plenoctrees} to encode the view direction, denote as $\texttt{SH}(\bm{\omega})$, before feeding into the neural shader.

\subsection{Improving the efficiency and quality via surface-guided volume sampling}
\label{subsec:sampling}

Volume rendering generates high-quality colors with no requirement of precise surface recovery \cite{mildenhall2020nerf}, but at the cost of dense point sampling along rays. Previous methods either uniformly sample points along a ray \cite{muller2022instant,wang2023neus2,li2022nerfacc}, or employ hierarchical sampling \cite{mildenhall2020nerf} that obtains points close to the surface in a multi-step fashion \cite{wang2021neus, yariv2021volume}; 
both strategies involve heavy computations and result in slow convergence.
% both of the strategies involve heavy computations and result in slow convergence.

With a maintained and synchronized surrogate $\widehat{\mathcal{S}}$ in Sur$^2$f, we expect it could be useful for guiding the point sampling in volume rendering, since for any ray $\bm{r}$, the ray-surface intersection can be efficiently and exactly obtained via rasterization or ray casting. Let $\bm{x}_{\widehat{\mathcal{S}}} = \bm{r}(t_{\widehat{\mathcal{S}}})$ be the intersection point, we conduct point sampling along $\bm{r}$ as 
\begin{equation}\label{eq:surfgs}
\begin{aligned}
\bm{x}_i = \bm{r}(t_i) = \bm{o} + t_i \bm{\omega} \quad \textrm{s.t.} \quad t_i \sim \mathcal{N}(t_{\widehat{\mathcal{S}}}, \sigma^2) , 
\end{aligned}
\end{equation}
where $\sigma^2$ is the variance that controls the tightness of the sampled points around $\bm{x}_{\widehat{\mathcal{S}}}$. We anneal $\sigma$ linearly w.r.t. the optimization iteration, such that the sampled points are distributed around the surface gradually.  
For each of those rays that do not intersect $\widehat{\mathcal{S}}$, we simply use uniform sampling along the ray.
%To further improve the efficiency, we also maintain a coarse dynamic occupancy grid, analogous to \cite{liang2023helixsurf}, that determines the empty space to remove points directly.
Volume sampling guided by $\widehat{\mathcal{S}}$ improves the efficiency greatly, as verified in Section \ref{sec:experiments}. 
% {\color{red} It also improves the quality of volume rendering, since XXX. Empirical results in Section \ref{sec:experiments} corroborate our analysis. }
It also improves the quality of volume rendering, since the learning is much concentrated on the surface. Empirical results in Section \ref{sec:experiments} corroborate our analysis.

\subsection{Training and Inference}
\label{subsec:training}

To train Sur$^2$f, in each iteration, we randomly sample pixels from the observed images and define the set of camera rays passing through these pixels as $\mathcal{R}$. By compactly writing $\bm{C}_v(\bm{r}; \bm{\theta}, \bm{\kappa}) = \sum_{i=1}^N T_i\alpha_i \bm{c}(\bm{x}_i, \bm{n}_i, \bm{F}_v(\bm{x}_i; \bm{\theta}), \bm{\omega}; \bm{\kappa})$ and $\bm{C}_s(\bm{r}; \bm{\vartheta}, \bm{\kappa}) = \bm{c}(\bm{x}_{\widehat{\mathcal{S}}}, \bm{n}_{\bm{x}_{\widehat{\mathcal{S}}}}, \bm{F}_s(\bm{x}_{\widehat{\mathcal{S}}}; \bm{\vartheta}), \bm{\omega}; \bm{\kappa})$, we train Sur$^2$f by minimizing the following loss:
\begin{equation}
\begin{aligned}
\mathcal{L}_s &= \frac{1}{\vert\mathcal{R}\vert} \sum_{\bm{r} \in \mathcal{R}} \texttt{L1}\left(\bm{C(r)}, \bm{C}_s(\bm{r}; \bm{\vartheta}, \bm{\kappa})\right) , \\
\mathcal{L}_v &= \frac{1}{\vert\mathcal{R}\vert} \sum_{\bm{r} \in \mathcal{R}} \texttt{Smooth}_\texttt{L1} \left(\bm{C(r)}, \bm{C}_v(\bm{r}; \bm{\theta}, \bm{\kappa})\right) , \\
  \mathcal{L} &= \lambda_s \mathcal{L}_s + \lambda_v \mathcal{L}_v +  \mathcal{L}_\text{\tiny Eik} ,
\end{aligned}
\end{equation}
where $\bm{C(r)}$ is the observed color for ray $\bm{r}$, and $\mathcal{L}_\text{\tiny Eik}$ is the Eikonal loss \cite{gropp2020implicit} that regularizes the learning of SDF $f$. $\lambda_s$ and $\lambda_v$ are hyperparameters weighting different loss terms. 
% In addition, we optimize an optional binary cross-entropy (BCE) loss once the image masks are available during the training process:
In addition, we optimize an optional binary cross-entropy (BCE) loss once the image masks are available:
\begin{equation}
    \mathcal{L}_\text{\tiny mask} = \frac{1}{\vert\mathcal{R}\vert} \sum_{\bm{r} \in \mathcal{R}} \mathds{1}_\text{\tiny mask}(\bm{r}) \texttt{BCE}\left(m(\bm{r}), O(\bm{r})\right),
\end{equation}
where $\mathds{1}_\text{\tiny mask}(\bm{r})$ indices whether the mask $m(\bm{r})$ of the ray is available, and $O (\bm{r}) = \sum^N_{i=1} T_i \alpha_i$ is the opacity of ray $\bm{r}$. The mask loss is weighted by the hyperparameter $\lambda_\text{\tiny mask}$.

During inference, we use marching cubes \cite{lorensen1987marching} to extract a surface mesh $\mathcal{S}$ from the learned $f$. Given the simultaneously learned neural shader, we are able to conduct real-time, interactive rendering of $\mathcal{S}$ with the mesh rasterized by an off-the-shelf graphics pipeline.

%-------------------------------------------------------------------------
\section{Experiments}
\label{sec:experiments}

In this section, we first verify our proposed Sur$^2$f in the context of surface reconstruction from multi-view images (\cf Section \ref{sec:method}) on the DTU object benchmark dataset \cite{jensen2014large} by comparing with baselines (\cf Section \ref{subsec:comparisons_dtu}) and designing elaborate experiments (\cf Section \ref{subsec:ablation}).
Then we conduct comparisons with other hybrid representation methods (\cf Section \ref{subsec:exp_cmphybrid}) in the task of physically based inverse rendering from multi-view images on the NeRF synthetic dataset \cite{mildenhall2020nerf}.
We further examine the applicability of Sur$^2$f by applying it to the task of indoor and outdoor scene reconstruction (\cf Section \ref{subsec:exp_indooroutdoor}) on ScanNet \cite{dai2017scannet} and Tanks and Temples \cite{knapitsch2017tanks} datasets.
We also apply Sur$^2$f to text-to-3D generation (as in \cref{fig:teaser}). More results are provided in the appendix.

\begin{figure*}
    \centering
    \resizebox{0.90\linewidth}{!}{\begin{tabular}{p{80pt}<{\centering}p{80pt}<{\centering}p{80pt}<{\centering}p{80pt}<{\centering}p{80pt}<{\centering}p{80pt}<{\centering}} Reference & \quad\quad NDS \cite{worchel2022multi} & \quad \ \ UNISURF \cite{oechsle2021unisurf} & \quad\quad NeuS \cite{wang2021neus} & \quad\quad\quad NeuS2 \cite{wang2023neus2} & \quad\quad\quad\quad Ours  \end{tabular}}
    \scalebox{0.94}{ 
    \includegraphics[width=\textwidth]{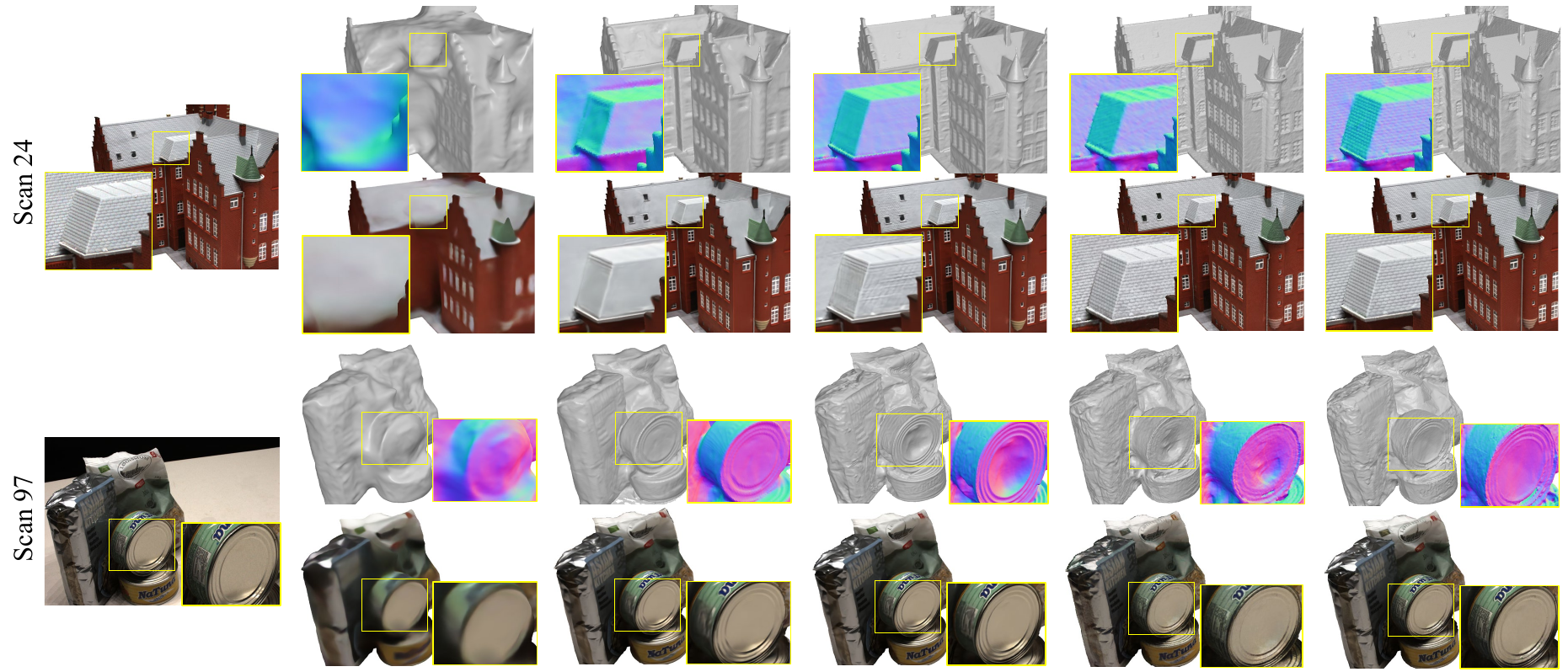}
    }
    \vspace{-0.3cm}
    \caption{
    \textbf{Qualitative comparisons on the DTU dataset \cite{jensen2014large}}. 
    The surface details are zoomed in and visualized as colors coded by surface normals for a better view. For each scene, the upper row is for geometry reconstruction while the lower row is for image synthesis.
    }
    \label{fig:comparison_dtu}
\vspace{-0.2cm}
\end{figure*}

\begin{table*}%[htb]
\centering
\scalebox{0.72}{ 
    \begin{tabular}{p{0.1cm}lcccccccccccccccc|c}
    \Xhline{3\arrayrulewidth}
    & Scan & 24   & 37   & 40   & 55   & 63   & 65   & 69   & 83   & 97   & 105  & 106  & 110  & 114  & 118  & 122  & Mean          & Time                                    \\ \cline{2-19}
    \multirow{4}{*}{\rotatebox[origin=c]{90}{CD $\downarrow$}} 
    & NDS \cite{worchel2022multi}            & 4.24          & 5.25          & 1.30          & 0.53          & 2.47          & 1.22          & 1.35          & 1.59          & 2.77          & 1.15          & 1.02          & 3.18          & 0.62          & 1.65          & 0.91          & 1.95          & \textit{7min}      \\
    & UNISURF \cite{oechsle2021unisurf}      & 1.32          & 1.36          & 1.72          & 0.44          & 1.35          & 0.79          & 0.80          & 1.49          & 1.37          & 0.89          & \textit{0.59} & 1.47          & 0.46          & 0.59          & 0.62          & 1.02          & 22h      \\
    & NeuS \cite{wang2021neus}               & 0.83          & 0.98          & 0.56          & \textit{0.37} & 1.13          & \textbf{0.59} & \textbf{0.60} & 1.45          & \textbf{0.95} & 0.78          & \textbf{0.52} & 1.43          & \textbf{0.36} & \textbf{0.45} & \textit{0.45} & 0.77          & 6.5h      \\
    & NeuS2 \cite{wang2023neus2}             & \textit{0.56} & \textit{0.76} & \textit{0.49} & \textit{0.37} & \textit{0.92} & 0.71          & 0.76          & \textit{1.22} & 1.08          & \textbf{0.63} & \textit{0.59} & \textit{0.89} & 0.40          & \textit{0.48} & 0.55          & \textit{0.70} & \textbf{5min}      \\\cline{2-19}
    & Ours                                   & \textbf{0.50} & \textbf{0.67} & \textbf{0.44} & \textbf{0.35} & \textbf{0.89} & \textit{0.62} & \textit{0.71} & \textbf{1.18} & \textit{1.06} & \textit{0.75} & \textit{0.59} & \textbf{0.78} & \textit{0.38} & 0.49          & \textbf{0.44} & \textbf{0.66} & \textbf{5min}      \\\hline\hline
    & Scan & 24   & 37   & 40   & 55   & 63   & 65   & 69   & 83   & 97   & 105  & 106  & 110  & 114  & 118  & 122  & Mean          & RT                                    \\ \cline{2-19}
    \multirow{4}{*}{\rotatebox[origin=c]{90}{PSNR $\uparrow$}} 
    & NDS \cite{worchel2022multi}            & 20.03          & 21.30          & 25.16          & 24.69          & 26.69          & 26.84          & 24.20          & 30.19          & 24.15          & 28.04          & 25.73          & 25.82          & 26.65          & 18.08          & 31.49          & 25.27          & \textbf{5ms}  \\
    & UNISURF \cite{oechsle2021unisurf}      & 26.98          & 25.34          & 27.30          & 27.62          & 32.76          & 32.77          & 29.85          & 34.15          & 28.84          & 32.50          & 33.47          & 31.45          & 29.60          & 35.49          & 35.76          & 30.93          & \textgreater{}1min \\
    & NeuS \cite{wang2021neus}               & 26.62          & 23.64          & 26.43          & 25.59          & 30.61          & 32.83          & 29.24          & 33.71          & 26.85          & 31.97          & 32.18          & 28.92          & 28.41          & 35.00          & 34.81          & 29.79          & 30s  \\
    & NeuS2 \cite{wang2023neus2}             & \textit{30.23} & \textit{27.29} & \textit{30.20} & \textit{33.27} & \textit{34.53} & \textit{33.29} & \textit{30.45} & \textit{37.73} & \textit{30.29} & \textit{34.26} & \textit{36.92} & \textit{34.39} & \textbf{33.50} & \textit{39.73} & \textit{40.30} & \textit{33.29} & \textit{1s}  \\\cline{2-19}
    & Ours                                   & \textbf{30.34} & \textbf{27.98} & \textbf{31.02} & \textbf{33.76} & \textbf{35.38} & \textbf{35.41} & \textbf{30.47} & \textbf{39.29} & \textbf{30.72} & \textbf{36.16} & \textbf{37.40} & \textbf{34.87} & \textit{32.74} & \textbf{39.82} & \textbf{40.53} & \textbf{33.95} & \textbf{5ms}  \\
    \Xhline{3\arrayrulewidth}
    \end{tabular}
}
\vspace{-0.3cm}
\caption{\textbf{Quantitative comparisons on the DTU dataset \cite{jensen2014large}}. We compare the reconstructed geometry with baseline methods using \emph{Chamfer Distance}. Our method achieves the best overall performance while performing as efficiently as NeuS2 \cite{wang2023neus2} and much faster than other methods.
For rendering PSNR comparisons, our method outperforms other methods in most cases for image synthesis quality with fast Rendering Time (RT).
We mark the \textbf{Best result} and the \textit{Second best result}.
%
% \textbf{Quantitative comparisons of reconstruction on the DTU dataset \cite{jensen2014large}}. We compare with the state-of-the-art multi-view surface reconstruction methods on both with (w/) and without (w/o) foreground mask settings using \emph{Chamfer Distance}. Our method achieves the best overall performance while performing as efficient as NeuS2 \cite{wang2023neus2} and much faster than other methods. \textbf{Bold} for the best result.
% \textbf{Rendering PSNR comparisons on the DTU dataset \cite{jensen2014large}.}
% Our method outperforms other methods in most cases for image synthesis quality with fast Rendering Time (RT).
% We mark the \textbf{Best result} and the \textit{Second best result}.
} 
\label{tab:dtu_cd}
\vspace{-0.5cm}
\end{table*}

\vspace{0.1cm}
\noindent{\textbf{Implementation Details}}
We implement Sur$^2$f in PyTorch\cite{paszke2019pytorch} framework with CUDA extensions.
We set the multi-resolution hash grids with 8 levels for the surface rendering branch and 14 levels for the volume rendering branch, where both hash resolutions range from 16 to 1024 with a feature dimension of 2, and set the spherical harmonics to 4 degree.
During training, we employ the Adam optimizer \cite{kingma2014adam} with a learning rate of 2e-3, and set the hyperparameters $\lambda_s$, $\lambda_v$, $\lambda_\text{\tiny Eik}$, and $\lambda_\text{\tiny mask}$ to 1, 1, 0.05, and 0.1, respectively. 
% In each iteration, we sample a dynamic batch of approximately 14800 rays and use customized CUDA kernels to calculate the $\alpha$-compositing colors of the sampled points along each ray as \cref{eq:volume_render}.
In each iteration, we sample a batch of 14800 rays, do ray casting with NVIDIA OptiX \cite{parker2010optix} engine, and use customized CUDA kernels to calculate the $\alpha$-compositing colors of the sampled points along each ray as \cref{eq:volume_render}.
% Furthermore, we apply a linear decrease in $\sigma$ with respect to the optimization iteration.
All experiments are conducted on a single GeForce RTX 3090 GPU. Further implementation details for different tasks can be found in the appendix.

\vspace{0.1cm}
\noindent{\textbf{Evaluation Metrics}}
For surface reconstruction, we measure the recovered surfaces with Chamfer Distance (CD) \cite{jensen2014large}. 
To evaluate the image rendering qualities, we compute the peak signal-to-noise ratio (PSNR) between the reference images and the synthesized images.
We compute the mean angle error (Mean-A) for the evaluation of normal maps.

\subsection{Comparisons on DTU in the context of multi-view surface reconstruction}
\label{subsec:comparisons_dtu}
We evaluate both the geometry reconstruction metric and the rendering quality of our proposed Sur$^2$f.
We set NDS \cite{worchel2022multi}, NeuS \cite{wang2021neus}, UNISURF \cite{oechsle2021unisurf}, and NeuS2 \cite{wang2023neus2} as our baselines. 
As NDS \cite{worchel2022multi} optimizes explicit mesh with a differentiable surface rendering paradigm, NeuS \cite{wang2021neus} takes the SDF-induced volume rendering paradigm that we based on, UNISURF \cite{oechsle2021unisurf} proposes to unify these two rendering paradigms, and NeuS2 \cite{wang2023neus2} serves as the speed baseline. 
The results are presented in \cref{tab:dtu_cd}.
In the realm of geometry reconstruction, Sur$^2$f achieves the best overall performance while maintaining the fastest convergence speed. 
% The qualitative results showcased in \cref{fig:comparison_dtu} further illustrate Sur$^2$f's ability to swiftly capture fine-grained geometry details without the need for a long-time optimization.
The qualitative results showcased in \cref{fig:comparison_dtu} further illustrate Sur$^2$f's ability to swiftly capture fine-grained geometry details.
Sur$^2$f also demonstrates superior image synthesis quality and achieves fast rendering in mere milliseconds.

\subsection{Ablation Studies}
\label{subsec:ablation}
% As stated in Section \ref{sec:method}, Sur$^2$f learns a synchronized pair of an implicit SDF and an explicit surrogate surface, and unifies volume rendering of the implicit SDF and surface rendering of the surrogate mesh with a shared neural shader. Sur$^2$f also improves efficiency and quality via surface-guided volume sampling.
% We design elaborate experiments to evaluate these designs on the DTU dataset \cite{jensen2014large}.
We design elaborate experiments to evaluate the designs of Sur$^2$f that are stated in Section \ref{sec:method}. These studies are conducted on the DTU dataset \cite{jensen2014large}.

\noindent{\textbf{Analysis on the synchronization of Surface Surrogate}}
Sur$^2$f maintains an explicit, surrogate surface mesh, which is synchronized by driving the deformation induced from the implicit SDF $f$ (\cf Section \ref{subsec:surrogate}). 
We visualize the surrogate deformation results and the distance between the surrogate mesh and the underlying surface represented by $f$ at certain training iterations in \cref{fig:ablfig_synsurrogate}, which shows that the surrogate mesh is synchronized with the learning of $f$ and they together converge to the same surface.
We also examine the intervals of iteration for applying isosurface extraction to re-boot the synchronization and various algorithms to perform the rebooting.
As can be seen in \cref{tab:abl_mcube}, we find that an interval of 500 iterations strikes a balance between accuracy and efficiency, and Sur$^2$f is not sensitive to the choice of isosurface extraction algorithms.

\begin{figure}%[htbp]
% \vspace{-0.1cm}
\centering
\scalebox{0.43}{
    \includegraphics[width=\textwidth]{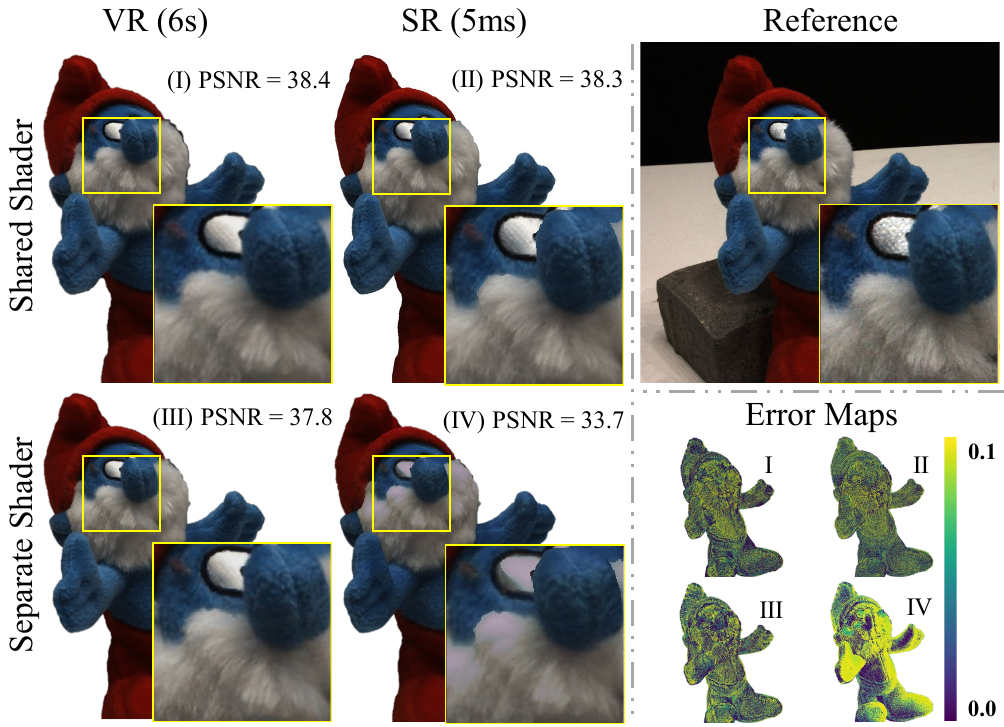}
}
\vspace{-0.15cm}
    \caption{Visualization of images synthesized by Volume Rendering (VR) and Surface Rendering (SR) with/without learning a shared shading network. The error maps measure the absolute pixel distance between rendered images and reference.
    }
    \label{fig:ablfig_alignvrsr}
\vspace{-0.2cm}
\end{figure}

\noindent{\textbf{Analysis on the shared shading network}}
We unify surface rendering and volume rendering with a shared shading network (\cf Section \ref{subsec:commonshading}), which promotes the parallelly learned surrogate mesh and implicit SDF to converge to the same surface, as evidenced by the previous ablation.
We further show the image synthesis results of the two rendering streams with/without learning a shared shader in \cref{fig:ablfig_alignvrsr}. 
% We find that the rendering results without a shared shader are much different, which reveals that the synchronization between surrogate mesh and implicit SDF is disturbed.   
We observe significant discrepancies in renderings without a shared shader, which suggests a disruption in the synchronization between the surrogate mesh and the implicit SDF.

\noindent{\textbf{Analysis on the surface-guided volume sampling strategy}} 
With a maintained and synchronized surface surrogate, Sur$^2$f further introduces a surface-guided volume sampling strategy to improve the sampling efficiency (\cf Section \ref{subsec:sampling}).
We compare this ray sampling strategy with alternative approaches, including uniform sampling proposed by Nerfacc \cite{li2022nerfacc} and hierarchical sampling adopted by NeuS \cite{wang2021neus}. The results in \cref{tab:abl_raysampling} show that our surface-guided sampling strategy leads to more accurate reconstruction and higher quality of rendering while saving training time.

\begin{table}%[]
    % \vspace{-0.3cm}
    \centering
    \scalebox{0.67}{
    \begin{tabular}{c|ccccc|cc}\hline
    \makecell{Algorithm\\Interval/iter}     & \makecell{MC\\1} & \makecell{MC\\300} & \makecell{MC*\\500} & \makecell{MC\\700} & \makecell{MC\\1000} & \makecell{DMTet\\500} & \makecell{FlexiCubes\\500} \\\hline
    CD $\downarrow$   & 0.654     & 0.659       & 0.661       & 0.682       & 0.715        & 0.662 & 0.661      \\
    Time/s $\downarrow$ & 21869     & 384         & 348         & 335         & 324          & 358   & 359 \\\hline     
    \end{tabular}
    }
    \vspace{-0.2cm}
    \caption{Analysis on different intervals of iteration and different isosurface extraction algorithms (\ie Marching Cube (MC) \cite{lorensen1987marching}, DMTet \cite{shen2021deep_dmtet} and FlexiCubes \cite{shen2023flexible}) for the retopology of surrogate.
    }
    \label{tab:abl_mcube}
    \vspace{-0.2cm}
\end{table}

\begin{table}%[]
% \vspace{-0.1cm}
\centering
\scalebox{0.7}{
\begin{tabular}{lccc}
\hline
                & Surface-guided* & Uniform     & Hierarchical \\\hline
CD $\downarrow$   & 0.661       & 0.732   & 0.662         \\
PSNR $\uparrow$   & 33.95       & 30.09   & 33.68         \\
Time/s $\downarrow$ & 348         & 337     & 9538          \\\hline
\end{tabular}
}
\vspace{-0.2cm}
\caption{
    Analysis of ray sampling strategies for volume rendering.
    % Analysis of different ray sampling strategies for volume rendering.
}
\label{tab:abl_raysampling}
\vspace{-0.5cm}
\end{table}

\begin{figure}%[htbp]
% \vspace{-0.5cm}
\centering
% \scalebox{0.45}{
%     \includegraphics[width=\textwidth]{images/main/expfig_pbr.pdf}
% }
% \rotatebox{90}{\resizebox{1.15\linewidth}{!}{\begin{tabular}{p{90pt}<{\centering}p{90pt}<{\centering}p{90pt}<{\centering}p{90pt}<{\centering}p{90pt}<{\centering}} \quad Applications & \makecell{\quad\quad\quad Ours \\ \quad\quad\quad 0.4h} & \makecell{\quad\quad FlexiCubes \cite{shen2023flexible} \\ \quad\quad1.8h} & \makecell{\quad\quad DMTet \cite{shen2021deep_dmtet} \\ \quad\quad 1.3h}  & \makecell{\quad\quad Reference \\ \quad\quad Time $\downarrow$} \end{tabular}}}
\rotatebox{90}{\resizebox{1.18\linewidth}{!}{
\begin{tabular}{p{95pt}<{\centering}p{10pt}<{\centering}p{100pt}<{\centering}p{5pt}<{\centering}p{140pt}<{\centering}p{110pt}<{\centering}p{110pt}<{\centering}} \quad\quad \Large Applications & ~ & \makecell{\quad\quad \Large Ours \large (0.4h)} & ~ &\makecell{\Large FlexiCubes \large \cite{shen2023flexible} (1.8h)} & \makecell{ \Large DMTet \large \cite{shen2021deep_dmtet} (1.3h)}  & \makecell{\Large Reference \large (Time $\downarrow$)}
\end{tabular}}}
\begin{minipage}[t]{0.87\linewidth}
    \centering
    \includegraphics[width=\textwidth]{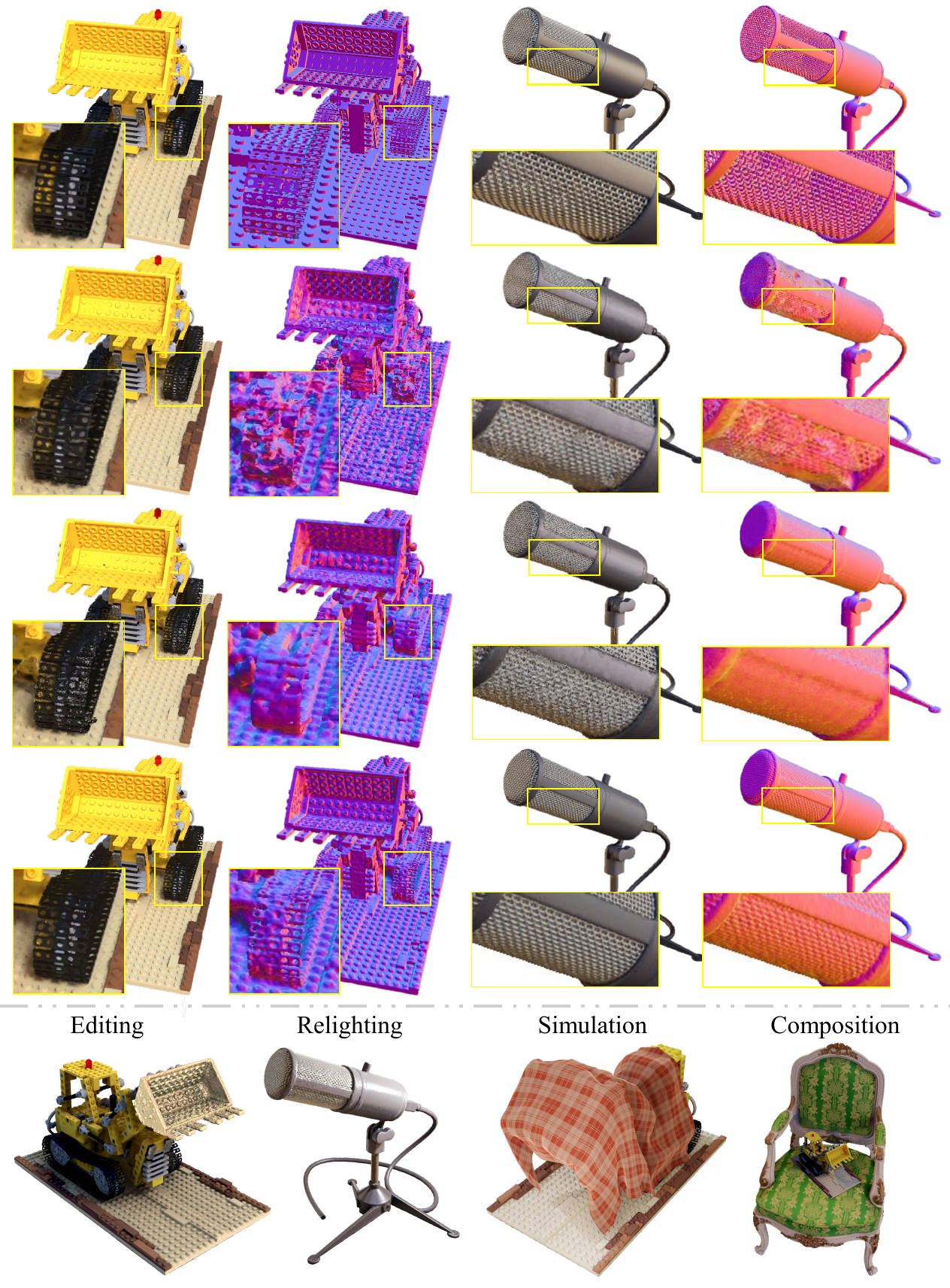}
\end{minipage}
\vspace{-0.25cm}
    \caption{
    Qualitative results on the NeRF synthetic dataset \cite{mildenhall2020nerf}.
    The last row shows some applications with a 3D content creation tool Blender \cite{blender}.
    }
    \label{fig:pbr}
\vspace{-0.4cm}
\end{figure}

\subsection{Comparison with other hybrid representations methods}
\label{subsec:exp_cmphybrid}
We compare our proposed Sur$^2$f with existing hybrid representations, \ie DMTet \cite{shen2021deep_dmtet} and FlexiCubes \cite{shen2023flexible}. 
Notably, these representations find application in NVDIFFREC \cite{nvdiffrec} for the task of physically based inverse rendering from multi-view images.
Thus we extend Sur$^2$f to the same task by incorporating the differentiable renderer from NVDIFFREC \cite{nvdiffrec}, and give the technical details in the appendix.
We compare the recovered normal maps and rendering results with these hybrid representations and report the quantitative and qualitative results in \cref{fig:pbr} and \cref{tab:pbr}, which demonstrate that Sur$^2$f promotes better quality and efficiency and also enables many downstream applications with an existing 3D content creation tool Blender \cite{blender}.

\begin{table}%[]
\vspace{-0.0cm}
\centering
\scalebox{0.7}{
\begin{tabular}{lcccccccc}
\hline
% \multirow{4}{*}{\rotatebox[origin=c]{90}{w/o mask}} 
PSNR $\uparrow$ & Chair & Drums & Ficus & Hotdog & Lego & Mats & Mic  & Ship \\\hline
DMTet      & 31.8  & 24.6  & 30.9  & 33.2   & 29.0 & 27.0 & 30.7 & 26.0 \\
FlexiCubes & 31.8  & 24.7  & 30.9  & 33.4   & 28.8 & 26.7 & 30.8 & 25.9 \\
Ours       & 31.9  & 25.2  & 31.4  & 33.9   & 30.3 & 27.0 & 31.3 & 27.1
\\\hline\hline
Mean-A $\downarrow$ & Chair & Drums & Ficus & Hotdog & Lego & Mats & Mic  & Ship \\\hline
DMTet      & 22.1  & 27.8  & 28.6  & 14.6   & 40.5 & 13.9 & 21.0 & 34.2 \\
FlexiCubes & 16.6  & 22.3  & 25.2  & 9.3    & 32.9 & 14.3 & 19.4 & 39.7 \\
Ours       & 14.0  & 20.0  & 22.1  & 8.6    & 29.2 & 12.0 & 16.3 & 22.7 \\\hline

\end{tabular}
}
\vspace{-0.2cm}
\caption{Quantitative results on the NeRF synthetic dataset \cite{mildenhall2020nerf}.
}
\label{tab:pbr}
\vspace{-0.2cm}
\end{table}

\subsection{Indoor and outdoor scene reconstruction}
\label{subsec:exp_indooroutdoor}

\begin{figure}%[htbp]
% \vspace{-0.5cm}
\centering
\resizebox{0.95\linewidth}{!}{\begin{tabular}{p{90pt}<{\centering}p{90pt}<{\centering}p{90pt}<{\centering}} Reference & Helixsurf \cite{liang2023helixsurf} & MonoSDF \cite{yu2022monosdf}  \end{tabular}}
\begin{minipage}{0.45\textwidth}
    \includegraphics[width=\textwidth]{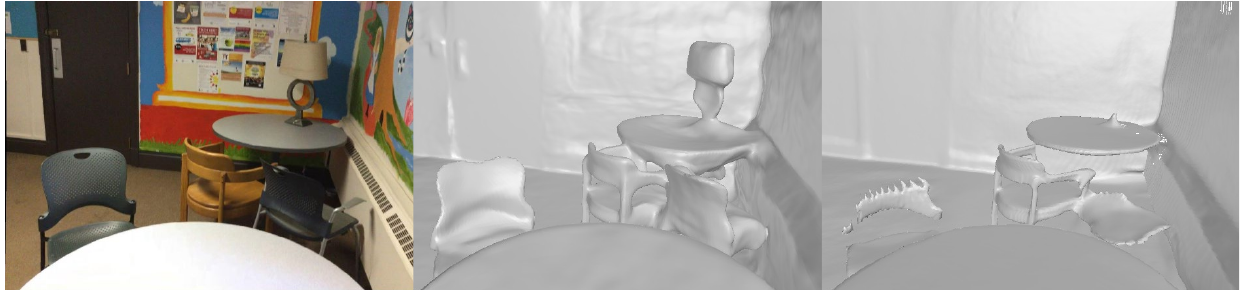}
\end{minipage}
\begin{minipage}{0.454\textwidth}
    \resizebox{1.0\linewidth}{!}{\begin{tabular}{p{90pt}<{\centering}p{90pt}<{\centering}p{90pt}<{\centering}} GT\quad & Helixsurf with Sur$^2$f & MonoSDF with Sur$^2$f  \end{tabular}}
    \includegraphics[width=\textwidth]{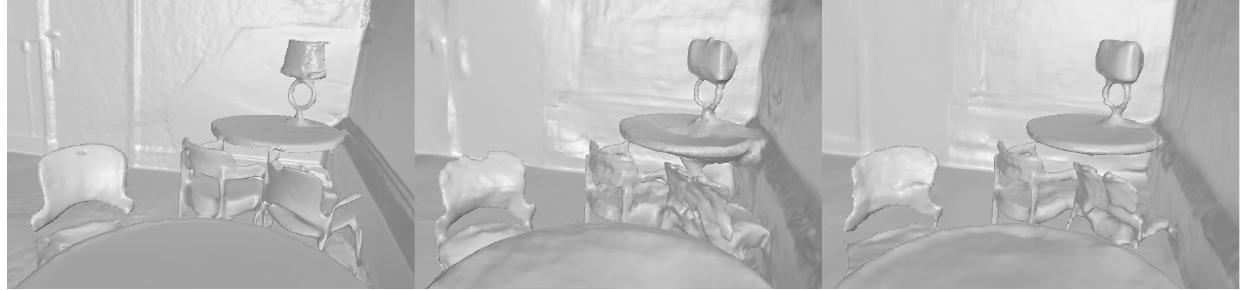}
\end{minipage}
\vspace{-0.2cm}
    \caption{
    \textbf{Qualitative results on ScanNet \cite{dai2017scannet}.}
    With Sur$^2$f, details (\eg the lamp, the chair back on the left) in the scene can be better reconstructed.
    }
    \label{fig:comparison_scannet}
\vspace{-0.2cm}
\end{figure}

\begin{figure}%[htbp]
% \vspace{-0.5cm}
\centering
\resizebox{1.0\linewidth}{!}{\begin{tabular}{p{130pt}<{\centering}p{140pt}<{\centering}p{110pt}<{\centering}} \quad\quad \Large Reference \large (Time $\downarrow$) & \Large Neuralangelo \large \cite{neuralangelo} (52h) & \Large Ours \large (1h)  \end{tabular}}
\scalebox{0.45}{
    \includegraphics[width=\textwidth]{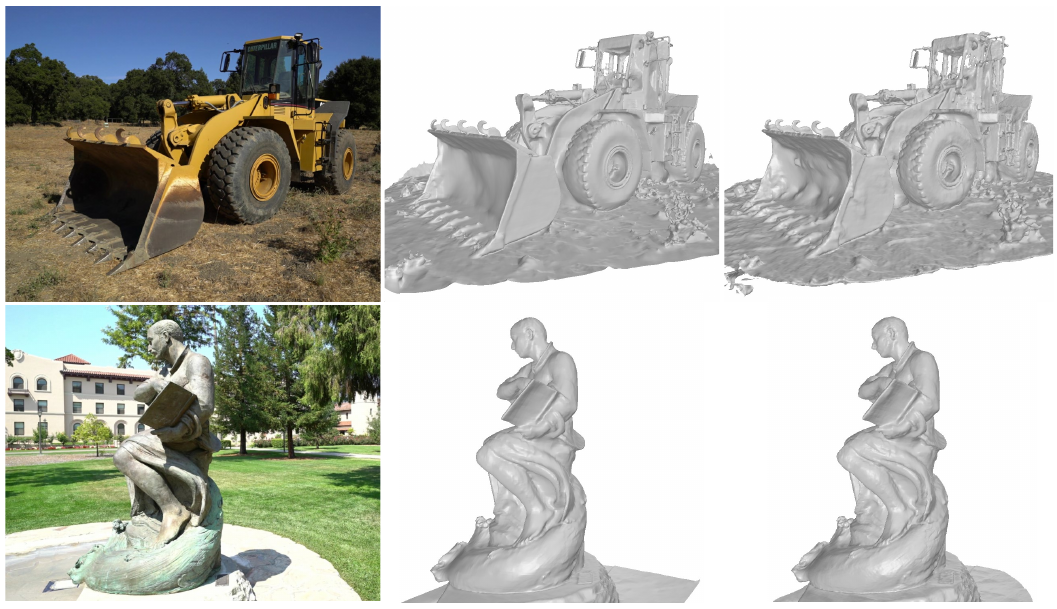}
}
\vspace{-0.2cm}
    \caption{
    \textbf{Qualitative results on Tanks and Temples \cite{knapitsch2017tanks}.} Sur$^2$f strikes a better quality-efficiency trade-off than Neuralangelo does.
    }
    \label{fig:comparison_tnt}
\vspace{-0.3cm}
\end{figure}

As a general hybrid representation, Sur$^2$f is also useful for indoor and large-scale outdoor reconstruction tasks. 
We incorporate Sur$^2$f with existing indoor scene reconstruction methods, \ie Helixsurf \cite{liang2023helixsurf} and MonoSDF \cite{yu2022monosdf}, to reconstruct the complex indoor scene from ScanNet \cite{dai2017scannet}.
And we reconstruct outdoor scenes from Tanks and Temples \cite{knapitsch2017tanks} by incorporating with Neuralangelo \cite{neuralangelo}. 
The results in \cref{fig:comparison_scannet} and \cref{fig:comparison_tnt} show that Sur$^2$f enhances the recovery of intricate scene details and gains a better quality-efficiency trade-off.

\section{Conclusion}
In this paper, we introduce a new hybrid representation, named Sur$^2$f.
By learning two parallel streams of an implicit SDF and an explicit surrogate surface mesh, and unifying volume rendering and surface rendering with a shared neural shader, 
Sur$^2$f exhibits outstanding performance across various surface modeling and reconstruction tasks.
Our approach reaffirms the efficacy of combining traditional rendering techniques with contemporary, differentiable rendering-based neural learning for robust surface modeling and 3D representation.

% We further improve the efficiency and quality via surface-guided volume sampling.
%-------------------------------------------------------------------------
% \section{Conclusion}
% In this paper, we introduce an efficient and high-quality multi-view surface reconstruction and rendering method, named Sur$^2$f.
% A surface surrogate is proposed that builds a bridge between surface rendering and volume rendering. We also propose a surface-guided volume sampling strategy and a common shader that accelerate the convergence of both renderings on reconstructing fine-grained details.
% Our approach reaffirms that combining traditional rendering techniques with recent, differentiable rendering based neural learning can be helpful for surface modeling and 3D representation.

%%%%%%%%%%%%%%%%%%%%%%%%%%%%%%%%%%%%%%%%%%%%%%%%%%%%%%%%%%%%%%%%%%%%%%%%%%%%%%%%%%%%%%%%%%%%%%%%%%%
% \clearpage
{
    \small
    \bibliographystyle{ieeenat_fullname}
    \bibliography{main}
}
%%%%%%%%%%%%%%%%%%%%%%%%%%%%%%%%%%%%%%%%%%%%%%%%%%%%%%%%%%%%%%%%%%%%%%%%%%%%%%%%%%%%%%%%%%%%%%%%%%%


%-------------------------------------------------------------------------
\clearpage

\maketitlesupplementary

\setcounter{section}{0}

% In this appendix, we provide supplementary details pertaining to Section \ref{sec:experiments}.
% To gain a more comprehensive understanding of the reconstructions and applications of Sur$^2$f, we encourage readers to refer to the accompanying supplementary video.

\section{Additional details for multi-view surface reconstruction}

\begin{figure*}[htbp]
\centering
\resizebox{0.90\linewidth}{!}{\begin{tabular}{p{80pt}<{\centering}p{80pt}<{\centering}p{80pt}<{\centering}p{80pt}<{\centering}p{80pt}<{\centering}p{80pt}<{\centering}} \quad\quad Reference & \quad\quad NDS \cite{worchel2022multi} & \quad \ \ UNISURF \cite{oechsle2021unisurf} & \quad\quad NeuS \cite{wang2021neus} & \quad\quad\quad NeuS2 \cite{wang2023neus2} & Sur$^2$f  \end{tabular}}
\scalebox{0.88}{
    \includegraphics[width=\textwidth]{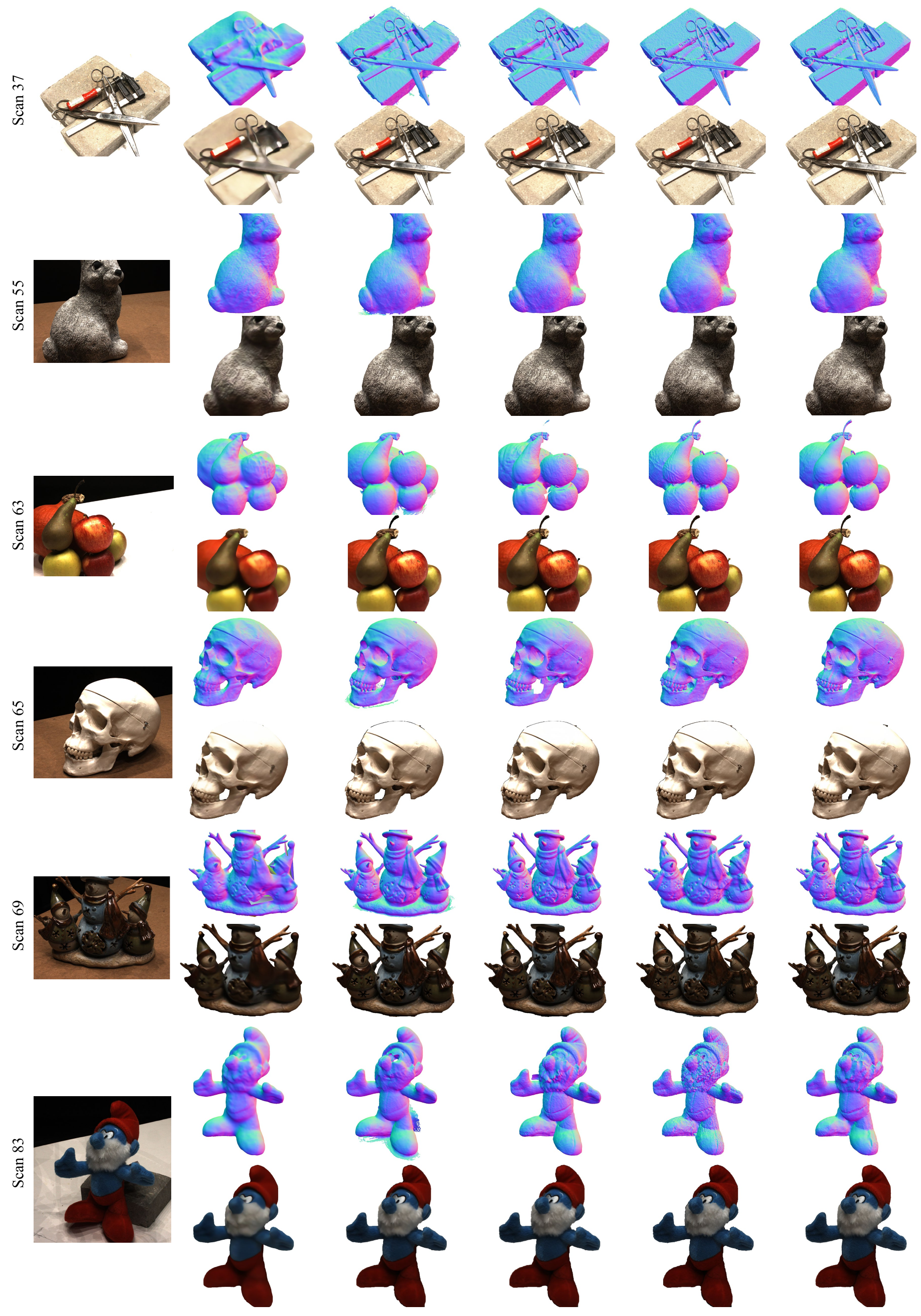}
}
\vspace{-0.1cm}
\caption{
    More qualitative results on the DTU dataset \cite{jensen2014large}. 
}
\label{fig:suppfig_dtup1}
\end{figure*}

\begin{figure*}[htbp]
\centering
\resizebox{0.90\linewidth}{!}{\begin{tabular}{p{80pt}<{\centering}p{80pt}<{\centering}p{80pt}<{\centering}p{80pt}<{\centering}p{80pt}<{\centering}p{80pt}<{\centering}} \quad\quad Reference & \quad\quad NDS \cite{worchel2022multi} & \quad \ \ UNISURF \cite{oechsle2021unisurf} & \quad\quad NeuS \cite{wang2021neus} & \quad\quad\quad NeuS2 \cite{wang2023neus2} & Sur$^2$f  \end{tabular}}
\scalebox{0.88}{
    \includegraphics[width=\textwidth]{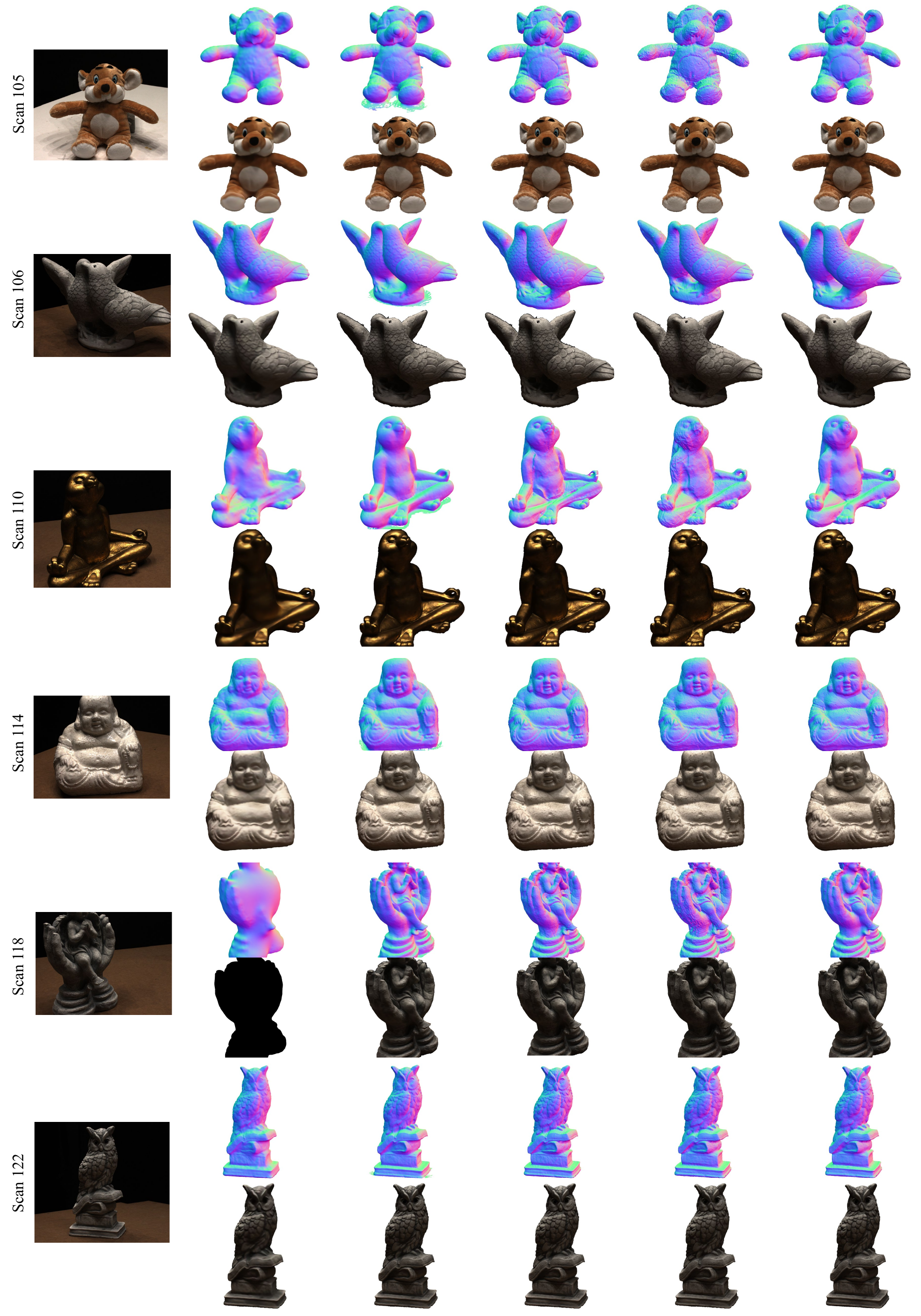}
}
\vspace{-0.1cm}
\caption{
    More qualitative results on the DTU dataset \cite{jensen2014large}.
}
\label{fig:suppfig_dtup2}
\end{figure*}

We use the same 15 object-centric scenes as in IDR \cite{yariv2020multiview} for quantitative and qualitative comparisons and analyses, where each scene provides 49 or 64 images with the resolution of 1600 $\times$ 1200 captured by a robot-held monocular RGB camera and ground truth point cloud obtained from a structured-lighted scanner. And the foreground masks are provided by IDR \cite{yariv2020multiview}. 
More qualitative results are shown in \cref{fig:suppfig_dtup1} and \cref{fig:suppfig_dtup2}.
% \hzj{More qualitative results are shown in XXX and XXX}.

\begin{figure}[htb]
\centering
\resizebox{0.95\linewidth}{!}{\begin{tabular}{p{70pt}<{\centering}p{70pt}<{\centering}p{70pt}<{\centering}} Reference & w/ mask & w/o mask  \end{tabular}}
\scalebox{0.46}{
    \includegraphics[width=\textwidth]{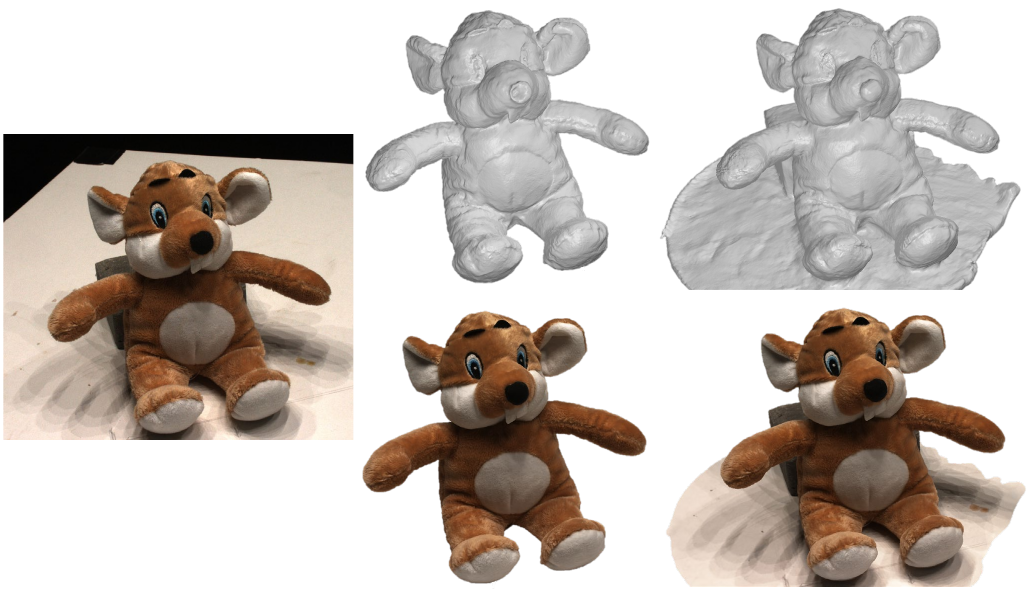}
}
\caption{Visualization of example reconstructions with (w/) and without (w/o) foreground mask.
}
\label{fig:ablfig_mask}
\end{figure}

Sur$^2$f also offers the flexibility to perform reconstruction without foreground masks.
If the foreground masks are not available, we model the foreground and background using two separate models by spatially separating the representations. 
The foreground area of interest for Sur$^2$f to reconstruct is defined as a sphere encompassing objects centered around an approximate scene center. 
Everything located outside the sphere is considered as background, which is represented with a NeRF model from NeRF++ \cite{zhang2020nerf++}.
Indeed, analogous strategies are adopted by NeuS \cite{wang2021neus} and Neuralangelo \cite{neuralangelo}.
We provide visual results of reconstructions for these two settings in \cref{fig:ablfig_mask}, which demonstrates that there is no noticeable difference in the reconstructions of the same foreground object.
Furthermore, we find the evaluation metric of \emph{Chamfer Distance} (w/ : 0.66 vs. w/o : 0.64) is close in these two settings as the evaluation is confined to the masked areas. 

\begin{figure}[htb]
\centering
\scalebox{0.48}{
    \includegraphics[width=\textwidth]{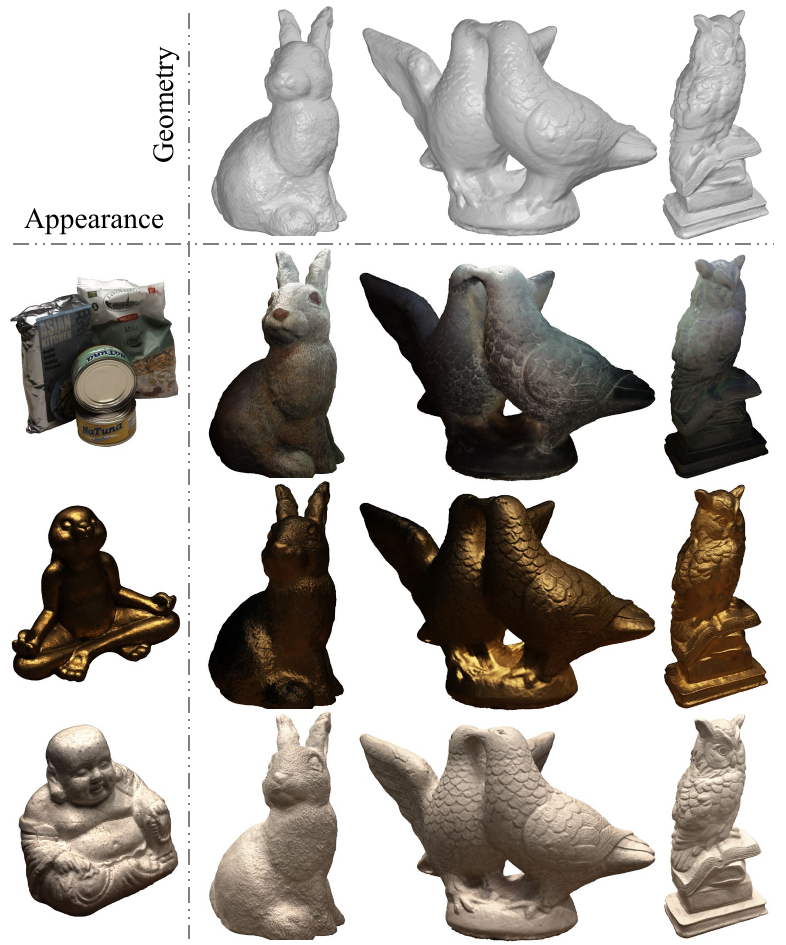}
}
\caption{Transferring different appearances to different geometries.
}
\label{fig:suppfig_dtu_trans}
\end{figure}

The framework design (illustrated in \cref{fig:pipeline}) of Sur$^2$f also facilitates the disentanglement between geometry and appearance \cite{yariv2020multiview}.
Leveraging this separation of geometry and appearance, different appearances can be seamlessly transferred to different geometries by exchanging the shading network (\cf Section \ref{subsec:commonshading}), as exemplified in \cref{fig:suppfig_dtu_trans}.

\section{Additional details for inverse rendering}
\begin{figure*}[htbp]
\centering
\resizebox{0.90\linewidth}{!}{\begin{tabular}{p{80pt}<{\centering}p{80pt}<{\centering}p{95pt}<{\centering}p{80pt}<{\centering}p{80pt}<{\centering}p{80pt}<{\centering}p{95pt}<{\centering}p{80pt}<{\centering}} \quad \Large Reference & \Large DMTet \cite{shen2021deep_dmtet} & \Large FlexiCubes \cite{shen2023flexible} & \Large Sur$^2$f & \quad 
 \Large  Reference & \Large DMTet \cite{shen2021deep_dmtet} & \Large FlexiCubes \large \cite{shen2023flexible} & \Large Sur$^2$f   \end{tabular}}
\scalebox{0.91}{
    \centering
    \includegraphics[width=\textwidth]{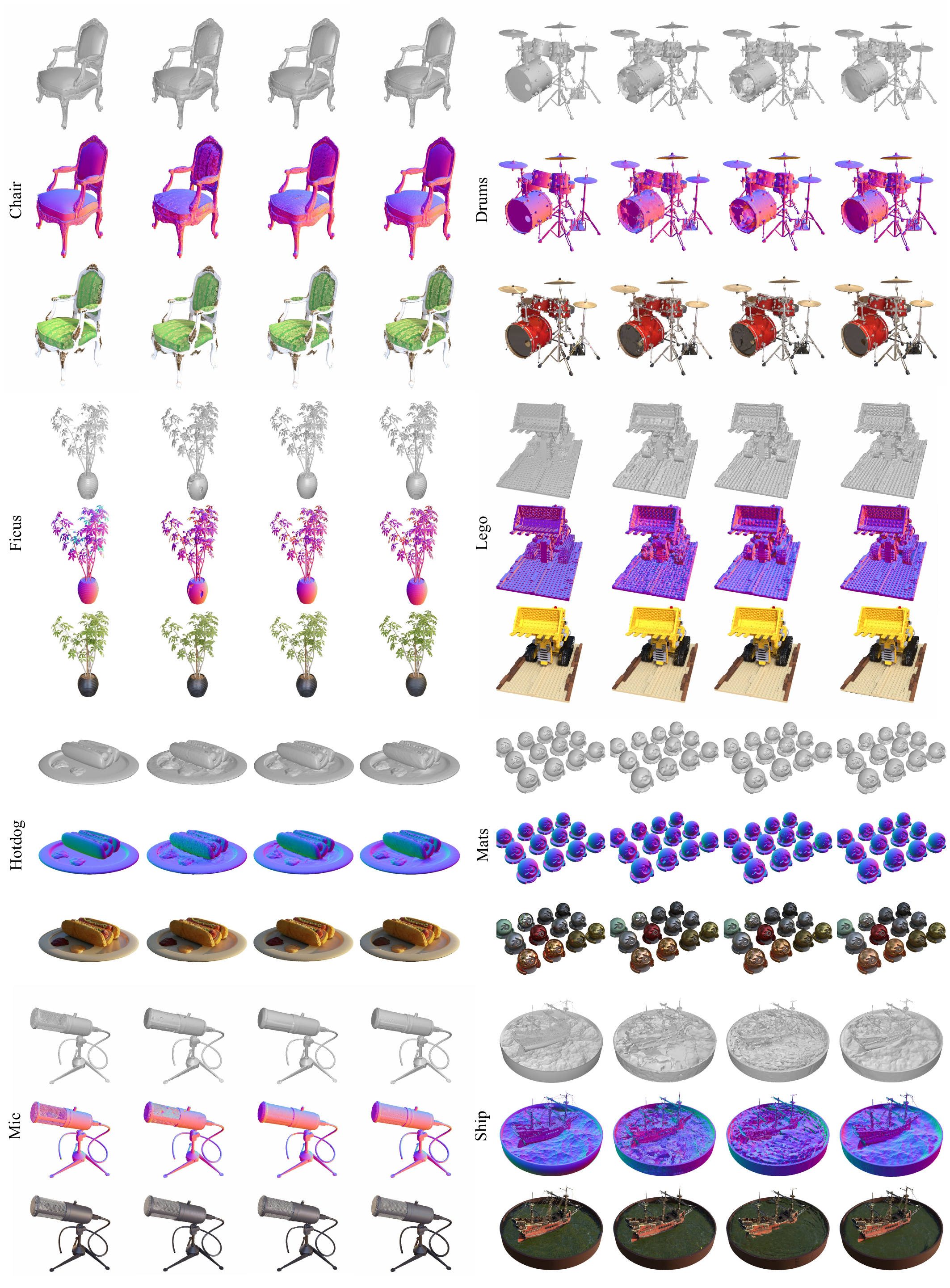}
}
\vspace{-0.1cm}
\caption{
    More qualitative results on the NeRF synthetic dataset \cite{mildenhall2020nerf}. 
}
\label{fig:suppfig_nerfsyn}
\end{figure*}

In Section \ref{subsec:exp_cmphybrid}, we compare Sur$^2$f with other hybrid representations in the task of physically based inverse rendering (PBR).
To achieve the inverse rendering from multi-view images, we incorporate Sur$^2$f with the differentiable render from NVDIFFREC \cite{nvdiffrec}. We give the details here and report more qualitative results in \cref{fig:suppfig_nerfsyn}.

\begin{figure}%[htb]
\centering
\scalebox{0.40}{
    \includegraphics[width=\textwidth]{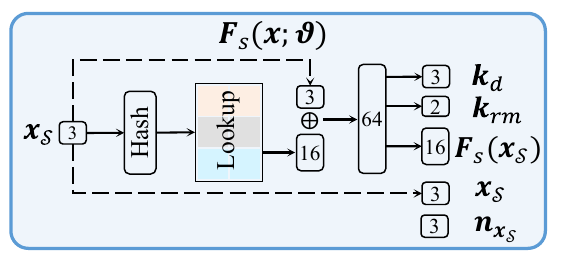}
}
\caption{
Adapted module for physically based inverse rendering. 
}
\label{fig:supp_pbrpipe}
\end{figure}

Starting from the non-emissive rendering equation \cref{eq:surface_rendering}, we use the Lambertian diffuse model \cite{basri2003lambertian} and the Cook-Torrance microfacet specular model \cite{cook1982reflectance} to formulate the BRDF $B(\bm{l}, -\bm{\omega})$:
\begin{equation}
\label{eq:brdf}
B(\bm{l}, -\bm{\omega}) = \underbrace{(1 - k_m) \frac{\bm{k}_d}{\pi}}_{\text{diffuse}} + \underbrace{\frac{
    DFG
}{
    4\left<\bm{n} \cdot \bm{l}\right>\left<\bm{n} \cdot -\bm{\omega}\right>
}}_{\text{specular}},
\end{equation}
where $\bm{k}_d \in [0, 1]^3$ and $k_m \in [0, 1]$ denote the base color and metallic value of surface point $\bm{x}_{\widehat{\mathcal{S}}}$ for the view-independent diffuse term. 
And for the view-dependent specular term, we define the surface roughness $k_r \in [0, 1]$ to condition the microfacet distribution function ($D$), Fresnel reflection ($F$), and geometric shadowing factor ($G$). We obtain $\bm{k}_d$ and $\bm{k}_{rm} = (k_r, k_m)$ by adapting the surface geometric feature learning module $\bm{F}_s(\bm{x}_{\widehat{\mathcal{S}}}; \bm{\vartheta})$, which is illustrated in \cref{fig:supp_pbrpipe}. 

According to \cref{eq:brdf}, the rendering equation is:
\begin{equation}
\label{eq:all_rendering}
\scriptsize
\begin{aligned}
L(\bm{x}_{\mathcal{S}}, \bm{\omega}) =& \int_\Omega L_i(\bm{x}_{\mathcal{S}}, \bm{l})B(\bm{l}, - \bm{\omega}) \left<\bm{l} \cdot \bm{n}_{\bm{x}_{\mathcal{S}}} \right> d\bm{l} \\
=& \int_\Omega L_i(\bm{x}_{\mathcal{S}}, \bm{l}) \left[
(1 - k_m) \frac{\bm{k}_d}{\pi} + \frac{
    DFG
}{
    4\left<\bm{n} \cdot \bm{l}\right>\left<\bm{n} \cdot -\bm{\omega}\right>
}
\right] \left<\bm{l} \cdot \bm{n}_{\bm{x}_{\mathcal{S}}} \right> d\bm{l} \\
=& (1 - k_m) \frac{\bm{k}_d}{\pi} \int_\Omega L_i(\bm{x}_{\mathcal{S}}, \bm{l}) \left<\bm{l} \cdot \bm{n}_{\bm{x}_{\mathcal{S}}} \right> d\bm{l} + \\
& \int_\Omega L_i(\bm{x}_{\mathcal{S}}, \bm{l}) \frac{
    DFG
}{
    4\left<\bm{n} \cdot \bm{l}\right>\left<\bm{n} \cdot -\bm{\omega}\right>
} \left<\bm{l} \cdot \bm{n}_{\bm{x}_{\mathcal{S}}} \right> d\bm{l} \\
=& L_\text{diffuse}(\bm{x}_{\mathcal{S}}) + L_\text{specular}(\bm{x}_{\mathcal{S}}, \bm{\omega}).
\end{aligned}
\end{equation}

\begin{figure*}%[htb]
\centering
\scalebox{0.99}{
    \includegraphics[width=\textwidth]{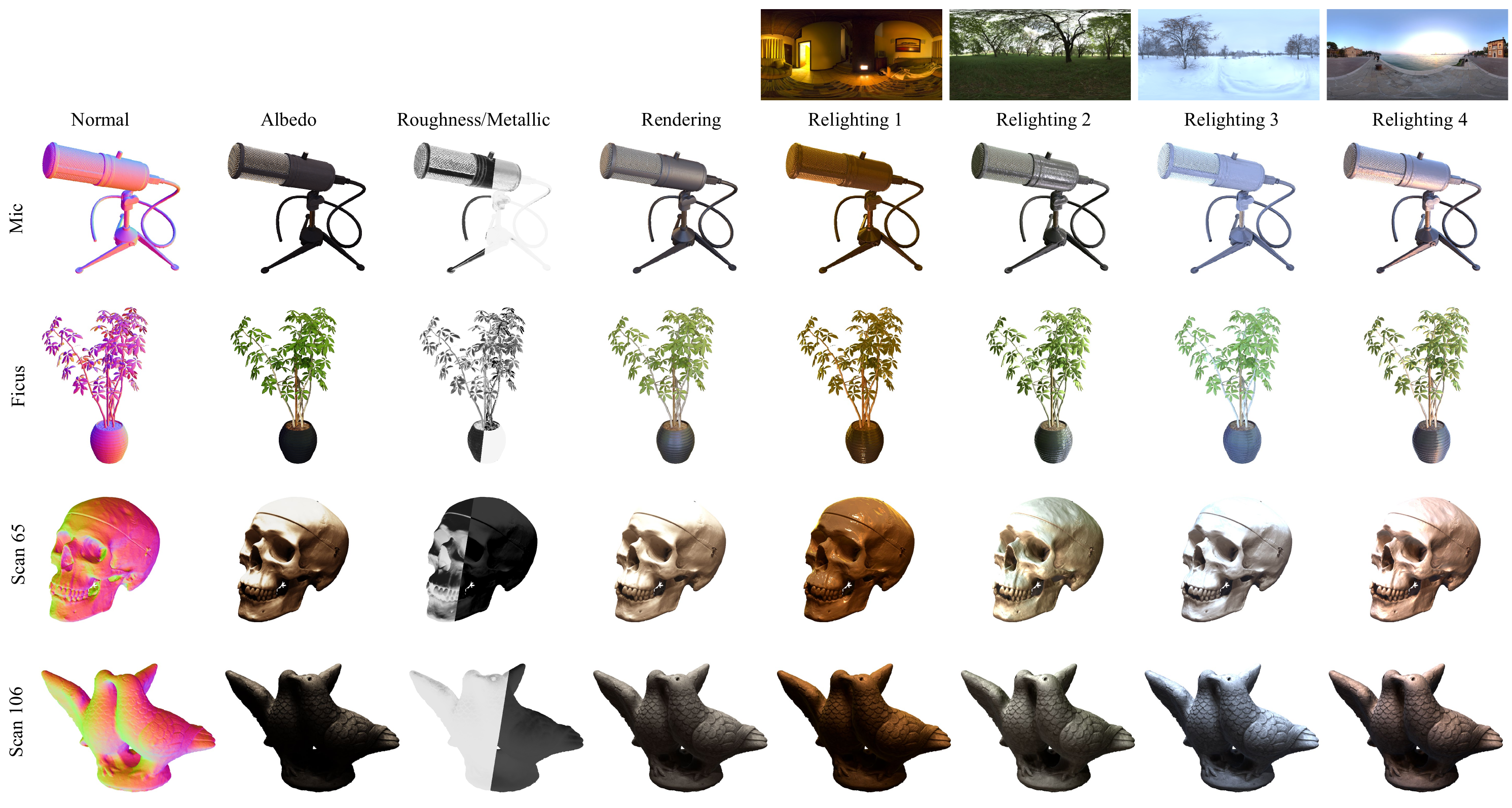}
}
\caption{
Reconstruction results and relighting application of physically based inverse rendering.
The objects of the first two rows are from the NeRF synthetic dataset \cite{mildenhall2020nerf}, and the last two rows are from the DTU dataset \cite{jensen2014large}.
}
\label{fig:suppfig_pbr_light}
\end{figure*}

In this work, we follow NVDIFFREC \cite{nvdiffrec} to use image-based lighting (IBL) to model the incident illumination, and employ split-sum approximation \cite{karis2013real} to tackle the intractable integral. Thus the incident illumination can be represented as a learnable environment map $\bm{E}$, and the diffuse component $L_\text{diffuse}(\bm{x}_{\mathcal{S}})$ is defined as:
\begin{equation}\small
\label{eq:diffuse}
\begin{aligned}
L_\text{diffuse}(\bm{x}_{\mathcal{S}}) =& (1 - k_m) \frac{\bm{k}_d}{\pi} \int_\Omega L_i(\bm{x}_{\mathcal{S}}, \bm{l}) \left<\bm{l} \cdot \bm{n}_{\bm{x}_{\mathcal{S}}} \right> d\bm{l} \\
\approx& (1 - k_m) \frac{\bm{k}_d}{\pi}\ \bm{I}_\text{d},
\end{aligned}
\end{equation}
where $\bm{I}_\text{d}$ denotes the diffuse irradiance obtained by integrating the upper hemisphere of the environment map $\bm{E}$. And for the specular component $ L_\text{specular}(\bm{x}_{\mathcal{S}}, \bm{\omega})$, we split the tricky integration into two trivial integrations and calculate these two integrals in advance:
\begin{equation}
\scriptsize
\label{eq:specular}
\begin{aligned}
& L_\text{specular}(\bm{x}_{\mathcal{S}}, \bm{\omega}) =
\int_\Omega L_i(\bm{x}_{\mathcal{S}}, \bm{l}) \frac{
    DFG
}{
    4\left<\bm{n} \cdot \bm{l}\right>\left<\bm{n} \cdot -\bm{\omega}\right>
}\left<\bm{l} \cdot \bm{n}_{\bm{x}_{\mathcal{S}}} \right> d\bm{l} \\
\approx&
\underbrace{
\int_\Omega \frac{
    DFG
}{
    4\left<\bm{n} \cdot \bm{l}\right>\left<\bm{n} \cdot -\bm{\omega}\right>
}\left<\bm{l} \cdot \bm{n}_{\bm{x}_{\mathcal{S}}} \right> d\bm{l}
}_{\text{Environment BRDF} - \bm{K}}
\underbrace{
\int_\Omega D L_i(\bm{x}_{\mathcal{S}}, \bm{l}) \left<\bm{l} \cdot \bm{n}_{\bm{x}_{\mathcal{S}}} \right> d\bm{l}
}_{\text{Pre-Filtered Environment Map} - \bm{I}_\text{s}},
\end{aligned}
\end{equation}
where $\bm{K}$ can be calculated in advance and stored in a 2D look-up texture, which can be search by roughness $k_r$ and $\left<\bm{n} \cdot -\bm{\omega}\right>$.
And we convolve the environment map $\bm{E}$ with multi-level GGX distribution \cite{walter2007microfacet} to get the remaining $\bm{I}_\text{s}$.
In summary, the approximated rendering equation is:
\begin{equation}
\label{eq:final_render}
L(\bm{x}_{\mathcal{S}}, \bm{\omega}) \approx (1 - k_m) \frac{\bm{k}_d}{\pi}\ \bm{I}_\text{d} + \bm{K} \bm{I}_\text{s}.
\end{equation}
Specifically, we integrate $\bm{I}_\text{d}$ and $\bm{I}_\text{s}$ from environment map $\bm{E}$ at the beginning of each training step, and get the rendered image color with standard linear-to-sRGB tone mapping \cite{srgb1996}: 
\begin{equation}
    \bm{C}_{\text{pbr}}(\bm{r}) = \texttt{ToneMap}(L(\bm{x}_{\mathcal{S}}, \bm{\omega})).
\end{equation}
The whole integration is differentiable to enable the additional loss optimization for PBR:
\begin{equation}
\mathcal{L}_{\text{pbr}} = \frac{1}{\vert\mathcal{R}\vert} \sum_{\bm{r} \in \mathcal{R}} \texttt{L1}\left(\bm{C(r)}, \bm{C}_{\text{pbr}}(\bm{r})\right).
\end{equation}
After optimization, we get the environment map $\bm{E}$ and the surface materials (\ie albedo $\bm{k}_d$, roughness $k_r$, and metallic value $k_m$) suitable for the downstream applications (\eg editing and relighting), and we show some examples in \cref{fig:suppfig_pbr_light}.

\section{Additional details for indoor scene reconstruction}

For indoor scene reconstruction (\cf Section \ref{subsec:exp_indooroutdoor}), we incorporate the method with iterative intertwined regularization (\ie Helixsurf \cite{liang2023helixsurf}) and the method with monocular geometric cues (\ie MonoSDF \cite{yu2022monosdf}) to address the challenge of rich textureless areas within the complex indoor scene.
Both of the methods rely on the supervision of depths and normals, hence, we derive Sur$^2$f to get the depth and normal for each ray $\bm{r}$ from \cref{eq:volume_render} as:

\begin{equation}
\label{eq:render_depthnorm}
    d(\bm{r}) = \sum^N_{i = 1} T_i \alpha_i t_i,
    \quad
    % \bm{n}(\bm{r}) = \nabla f(\bm{o} + d(\bm{r}) \bm{\omega}) ,
    \bm{n}(\bm{r}) = \sum^N_{i = 1} T_i \alpha_i \bm{n}_i.
\end{equation}

\noindent
With the proposed surrogate surface, we check the multi-view geometric consistency and give the losses as:
\begin{equation} \small
\label{eq:loss_depthnorm}
\begin{aligned}
    & \mathcal{L}_\text{\tiny depth} = \frac{1}{\vert\mathcal{R}\vert} \sum_{\bm{r} \in \mathcal{R}}  \Gamma \left(0.05, \left\Vert \widetilde{d(\bm{r})} - t_{\widehat{\mathcal{S}}}\right\Vert_1\right) \texttt{L1}\left(\widetilde{d(\bm{r})}, d(\bm{r})\right) , \\
    & \mathcal{L}\text{\tiny normal} = \frac{1}{\vert\mathcal{R}\vert} \sum_{\bm{r} \in \mathcal{R}}  \Gamma \left(\left< \widetilde{\bm{n}(\bm{r})} \cdot \bm{n}_{\bm{x}_{\widehat{\mathcal{S}}}} \right>,  0.5 \right) \texttt{L1}\left(\widetilde{\bm{n}(\bm{r})}, \bm{n}(\bm{r}) \right), \\
    & \Gamma(x, y) = \left\{ \begin{array}{l}
        1 \quad \text{if } x \geq y \\
        0 \quad \text{if } x < y,
        \end{array}\right.
\end{aligned}
\end{equation}
where $\widetilde{d(\bm{r})}$ and $\widetilde{\bm{n}(\bm{r})}$ are the depth and normal supervision provided by Helixsurf \cite{liang2023helixsurf} or MonoSDF \cite{yu2022monosdf}, $\left<\cdot\right>$ is a descriptor that computes the cosine of two input vectors, $t_{\widehat{\mathcal{S}}}$ is given near \cref{eq:surfgs}.

% To further examine the applicability and scalability of Sur$^2$f, we conduct reconstruction on a complex indoor scene from ScanNet \cite{dai2017scannet}. 
% In addressing the challenge of rich textureless areas within the scene, we incorporate the iterative intertwined regularization method from Helixsurf \cite{liang2023helixsurf} and utilize monocular geometric cues from MonoSDF \cite{yu2022monosdf}, respectively. As illustrated in \cref{fig:comparison_scannet}, our Sur$^2$f extension significantly enhances the recovery of intricate scene details. 
% And we further test our methods on large-scale outdoor scene \cite{knapitsch2017tanks} reconstruction as shown in \cref{fig:comparison_tnt}, where Sur$^2$f shows comparable results with much higher efficiency.

% The depth $d$ of the surface from the camera center $\bm{o}$ can be approximated along the ray $\bm{r}$ as well, giving rise to
% \begin{equation}
% \label{eq:render_depthnorm}
%     d(\bm{r}) = \sum^N_{i = 1} T_i \alpha_i t_i,
%     \quad
%     \bm{n}(\bm{r}) = \nabla f(\bm{o} + d(\bm{r}) \bm{\omega}) ,
% \end{equation}
% where $\bm{n}(\bm{r}) \in \mathbb{R}^3$ denotes the surface normal at the intersection point and $\nabla f(\bm{x})$ is the gradient of SDF at $\bm{x}$.

\section{Additional details for text-to-3D generation}

\def\mathbi#1{\textbf{\em \text{#1}}}
\def\mathi#1{\textit{\text{#1}}}

\begin{figure*}[htbp]
\centering
% \resizebox{0.2\linewidth}{!}{\textit{A blue tulip}}
\resizebox{0.2\linewidth}{!}{\begin{tabular}{p{90pt}<{\centering}} \textit{A blue tulip}  \end{tabular}}
\begin{minipage}{1.0\textwidth}
    \includegraphics[width=\textwidth]{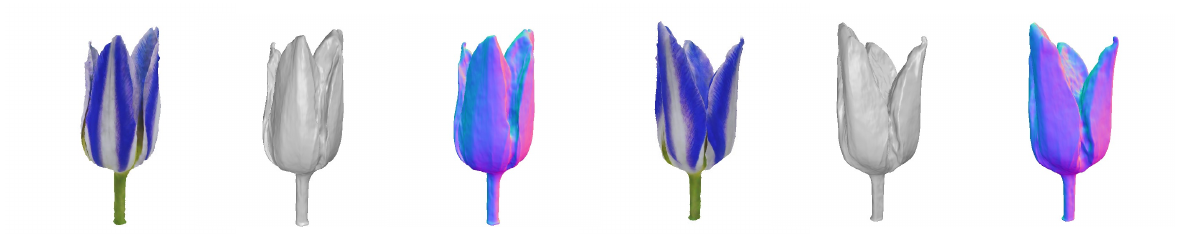}
\end{minipage}
\vspace{0.1cm}
\resizebox{0.2\linewidth}{!}{\begin{tabular}{p{90pt}<{\centering}} \textit{A delicious croissant}  \end{tabular}}
\begin{minipage}{1.0\textwidth}
% \resizebox{0.2\linewidth}{!}{\textit{A delicious croissant}}
    \includegraphics[width=\textwidth]{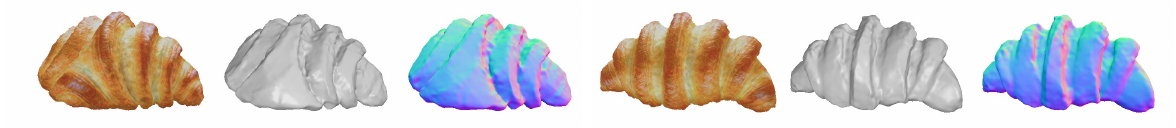}
\end{minipage}
\vspace{0.1cm}
% \resizebox{0.2\linewidth}{!}{\textit{A pineapple}}
\resizebox{0.2\linewidth}{!}{\begin{tabular}{p{90pt}<{\centering}} \textit{A pineapple}  \end{tabular}}
\begin{minipage}{1.0\textwidth}
    \includegraphics[width=\textwidth]{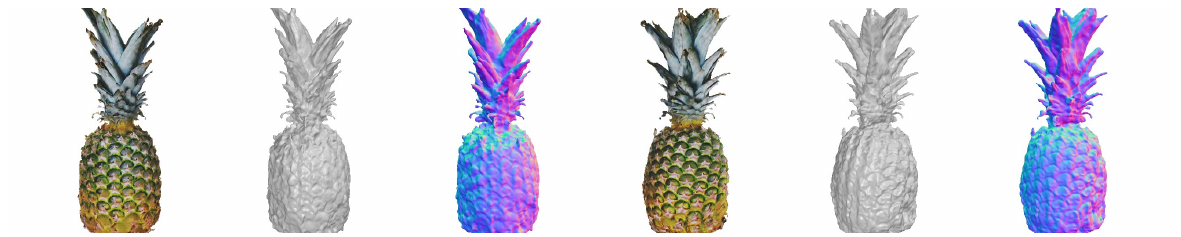}
\end{minipage}
    \caption{
        Examples of Text-to-3D Generation.
    }
    \label{fig:supp_tto3d}
\end{figure*}

For text-to-3D generation, we use the latent space diffusion model of Stable Diffusion \cite{rombach2022high} as our guidance. 
To update the parameters $\bm{\theta},\bm{\vartheta}, \bm{\kappa}$ in Sur$^2$f, we calculate the gradient by the Score Distillation Sampling (SDS) losses \cite{poole2022dreamfusion}:
\begin{equation} \small
\begin{aligned}
    \nabla_{(\bm{\theta}, \bm{\kappa})}&\mathcal{L}_{\text{\tiny SDS}} (\phi, \bm{\mathsf{x}}=\{\bm{C}_v(\bm{r}; \bm{\theta}, \bm{\kappa})\}) \\ &= \mathbb{E}\left[ \mathi{w}(\mathsf{t}) \left( \hat{\epsilon}_{\phi}(\bm{\mathsf{z}}_{\mathsf{t}}^{\bm{\mathsf{x}}}; \mathsf{y}, \mathsf{t}) - \epsilon \right)\frac{\partial \bm{\mathsf{x}}}{\partial (\bm{\theta}, \bm{\kappa})} \frac{\partial \bm{\mathsf{z}}^{\bm{\mathsf{x}}}}{\partial \bm{\mathsf{x}}} \right], \\
    \nabla_{(\bm{\vartheta}, \bm{\kappa})}&\mathcal{L}_{\text{\tiny SDS}} (\phi, \bm{\mathsf{x}}=\{\bm{C}_s(\bm{r}; \bm{\vartheta}, \bm{\kappa})\}) \\ &= \mathbb{E}\left[ \mathi{w}(\mathsf{t}) \left( \hat{\epsilon}_{\phi}(\bm{\mathsf{z}}_{\mathsf{t}}^{\bm{\mathsf{x}}}; \mathsf{y}, \mathsf{t}) - \epsilon \right)\frac{\partial \bm{\mathsf{x}}}{\partial (\bm{\vartheta}, \bm{\kappa})} \frac{\partial \bm{\mathsf{z}}^{\bm{\mathsf{x}}}}{\partial \bm{\mathsf{x}}} \right],
\end{aligned}
\end{equation}
where the rendered image $\bm{\mathsf{x}}$ connects with the image encoder of the pre-trained stable diffusion model that is parameterized as $\phi$, $\bm{\mathsf{z}}^{\bm{\mathsf{x}}}$ is the encoded latent code, $\hat{\epsilon}_{\phi}(\bm{\mathsf{z}}_{\mathsf{t}}^{\bm{\mathsf{x}}}; \mathsf{y}, \mathsf{t})$ is the predicted noise given text embedding $\mathsf{y}$ and timestep $\mathsf{t}$, and $\epsilon$ is the noise added in $\bm{\mathsf{z}}^{\bm{\mathsf{x}}}$; and $\mathi{w}(\mathsf{t})$ is a weighting function.
We give some generated examples in \cref{fig:supp_tto3d}.

% \section{Limitations}
% Although Sur$^2$f 

\end{document}